%% file: main.tex
\documentclass{article}

\usepackage[final]{neurips_2025}

\input{preamble}

\input{macros}

\usepackage{listings}
\usepackage[utf8]{inputenc} %
\usepackage[T1]{fontenc}    %
\usepackage{url}            %
\usepackage{booktabs}       %
\usepackage{amsfonts}       %
\usepackage{nicefrac}       %
\usepackage{microtype}      %
\usepackage{xcolor}         %
\usepackage{thm-restate}
\usepackage{siunitx}

\hypersetup{
	colorlinks=true,       %
	linkcolor=blue,        %
	citecolor=blue,        %
	filecolor=magenta,     %
	urlcolor=blue         
}

\definecolor{codegreen}{rgb}{0,0.6,0}
\definecolor{codegray}{rgb}{0.5,0.5,0.5}
\definecolor{codepurple}{rgb}{0.58,0,0.82}
\definecolor{backcolour}{rgb}{0.95,0.95,0.95}

\lstdefinestyle{mystyle}{
    backgroundcolor=\color{backcolour},   
    commentstyle=\color{codegreen},
    keywordstyle=\color{magenta},
    numberstyle=\tiny\color{codegray},
    stringstyle=\color{codepurple},
    basicstyle=\ttfamily\footnotesize,
    breakatwhitespace=false,         
    breaklines=true,                 
    captionpos=b,                    
    keepspaces=true,                 
    numbers=left,                    
    numbersep=5pt,                  
    showspaces=false,                
    showstringspaces=false,
    showtabs=false,                  
    tabsize=2,
    frame=single, %
    rulecolor=\color{black}
}

\title{Progressive Inference-Time Annealing of Diffusion Models for Sampling from Boltzmann Densities}

\author{%
Tara Akhound-Sadegh$^{1,2}$\thanks{Equal contribution $^{\dagger}$Equal advising. Correspondence to \texttt{tara.akhoundsadegh@mila.quebec, k.necludov@gmail.com, atong@aithyra.at}},
Jungyoon Lee$^{3,2}$\footnotemark[1],
Avishek Joey Bose$^{4,2}$,
Valentin De Bortoli$^{5}$, \\
\textbf{
Arnaud Doucet$^{5}$,
Michael M. Bronstein$^{4,6}$,
Dominique Beaini$^{7,3,2}$,
Siamak Ravanbakhsh$^{1,2}$,
} \\
\textbf{
Kirill Neklyudov$^{3,2,8 \dagger}$,
Alexander Tong$^{6,3,2 \dagger}$
}\\
$^1$McGill University, $^2$Mila -- Quebec AI Institute,  $^3$Université de Montréal,  $^4$University of Oxford,  \\ $^5$Google DeepMind, $^6$AITHYRA, $^7$Valence Labs, $^8$Institut Courtois
}

\begin{document}
\maketitle

\begin{abstract}
\looseness=-1
Sampling efficiently from a target unnormalized probability density remains a core challenge, with relevance across countless high-impact scientific applications. 
A promising approach towards this challenge is the design of amortized samplers that borrow key ideas, such as probability path design, from state-of-the-art generative diffusion models. However, all existing diffusion-based samplers remain unable to draw samples from distributions at the scale of even simple molecular systems.
In this paper, we propose \namelong (\nameshort) a novel framework to learn diffusion-based samplers that combines two complementary interpolation techniques: I.) Annealing of the Boltzmann distribution and II.) Diffusion smoothing. \nameshort trains a sequence of diffusion models from high to low temperatures by sequentially training each model at progressively lower temperatures, leveraging engineered easy access to samples of the temperature-annealed target density.
In the subsequent step, \nameshort enables simulating the trained diffusion model to \emph{procure training samples at a lower temperature} for the next diffusion model through inference-time annealing using a novel Feynman-Kac PDE combined with Sequential Monte Carlo. 
Empirically, \nameshort enables, for the first time, equilibrium sampling of $N$-body particle systems, Alanine Dipeptide, and Tripeptide in Cartesian coordinates with dramatically fewer energy function evaluations. Code available at: \url{https://github.com/taraak/pita}. 
\end{abstract}
\input{text/intro}

\input{text/background}
\input{text/method}

\input{text/related_work}

\input{text/experiments}

\input{text/conclusion}
\clearpage
\input{text/acknowledgements}
\bibliographystyle{apalike}
\bibliography{refs}
\clearpage
\clearpage
\appendix
\input{text/appendix}

\end{document}

%% file: preamble.tex
\usepackage{silence}
\usepackage[utf8]{inputenc}
\usepackage[T1]{fontenc}
\usepackage{tabularx}
\usepackage[final,pagebackref=true]{hyperref} %

\usepackage{lipsum}
\usepackage{url}
\usepackage{placeins}
\usepackage{booktabs}
\usepackage{soul}
\usepackage{amsfonts}
\usepackage{nicefrac}
\usepackage{microtype}
\usepackage{natbib}
\usepackage{proba}
\usepackage{wrapfig}
\usepackage{caption}
\usepackage{amssymb}
\usepackage{pifont}

\usepackage{subcaption}
\usepackage{float}
\usepackage{rotating}
\usepackage{etoolbox}
\usepackage{xparse}
\usepackage{grffile}
\usepackage{amsmath}
\usepackage{xspace}
\usepackage{euscript}
\usepackage{enumitem}
\usepackage{mathtools}
\usepackage{tikz}
\usepackage{tablefootnote}
\usepackage[flushleft]{threeparttable}
\usepackage{multirow}
\usepackage[title]{appendix}
\usepackage{arydshln}
\usepackage{array, booktabs}
\usepackage{comment}
\usepackage{graphicx}
\usepackage{adjustbox}
\usepackage{dsfont}
\usepackage{amsmath,amsthm}
\usepackage{balance}
\usepackage{multicol}
\usepackage{xcolor}
\usepackage{nameref}
\usepackage{pdfpages}
\usepackage{braket}
\usepackage[normalem]{ulem}
\usepackage{bbding}
\usepackage{xfrac}
\usepackage[usestackEOL]{stackengine}
\usepackage{indentfirst}
\usepackage{csquotes}
\usepackage{pgf}
\usepackage{varwidth}
\usepackage{verbatimbox}
\usepackage{verbatim}
\usepackage{bm}
\usepackage{algorithm, algorithmic}
\usepackage[colorinlistoftodos,prependcaption,textsize=small]{todonotes}
\usepackage{wrapfig} %
\usepackage[capitalise]{cleveref}   
\usepackage{empheq}
\definecolor{myblue}{rgb}{.8, .8, 1}

\definecolor{pastelblue}{RGB}{76,113,175}
\definecolor{pastelgreen}{RGB}{144,238,144}
\definecolor{pastelred}{RGB}{196,78,82}
\definecolor{pastelgrey}{RGB}{230,230,230}
\definecolor{pastelbeige}{RGB}{243,236,221}
\definecolor{pastelpurple}{RGB}{154,139,192}
\definecolor{salmon}{RGB}{250, 128, 114}
\definecolor{darkgreen}{rgb}{0,0.6,0}
\definecolor{darkred}{rgb}{0.5,0,0}
\definecolor{verylightgreen}{HTML}{F6FFF9}
\definecolor{verylightred}{HTML}{FFF4F3}
\definecolor{verylightgray}{HTML}{F4F6F6}
\definecolor{babyblueeyes}{rgb}{0.63, 0.79, 0.95}
\definecolor{lightpink}{rgb}{1.00, 0.714, 0.757}

\usepackage{tikzpeople}
\usetikzlibrary{shapes,decorations,arrows,calc,arrows.meta,fit,positioning}

\tikzset{
    -Latex,auto,node distance =1 cm and 1 cm,semithick,
    state/.style ={ellipse, draw, minimum width = 0.7 cm},
    point/.style = {circle, draw, inner sep=0.04cm,fill,node contents={}},
    bidirected/.style={Latex-Latex,dashed},
    el/.style = {inner sep=2pt, align=left, sloped}
}

\makeatletter
\def\thmt@refnamewithcomma #1#2#3,#4,#5\@nil{%
	\@xa\def\csname\thmt@envname #1utorefname\endcsname{#3}%
	\ifcsname #2refname\endcsname
	\csname #2refname\expandafter\endcsname\expandafter{\thmt@envname}{#3}{#4}%
	\fi}
\makeatother
\Crefname{conjecture}{Conjecture}{Conjectures}
\Crefname{definition}{Definition}{Definitions}
\Crefname{observation}{Observation}{Observations}
\Crefname{assumption}{Assumption}{Assumptions}
\Crefname{axiom}{Axiom}{Axioms}
\Crefname{case}{Case}{Cases}
\Crefname{claim}{Claim}{Claims}
\Crefname{conclusion}{Conclusion}{Conclusions}
\Crefname{condition}{Condition}{Conditions}
\Crefname{criterion}{Criterion}{Criteria}
\Crefname{exercise}{Exercise}{Exercises}
\Crefname{example}{Example}{Examples}
\Crefname{notation}{Notation}{Notations}
\Crefname{problem}{Problem}{Problems}
\Crefname{property}{Property}{Properties}
\Crefname{remark}{Remark}{Remarks}
\Crefname{solution}{Solution}{Solutions}
\Crefname{summary}{Summary}{Summaries}
\Crefname{motivation}{Motivation}{Motivations}
\Crefname{query}{Query}{Queries}

\newcommand{\one}{\mathds{1}}

\newcommand*\dbar[1]{\overline{\overline{\lower0.2ex\hbox{$#1$}}}}

\def\cE{{\mathcal{E}}}

\DeclareFontFamily{U}{BOONDOX-calo}{\skewchar\font=45 }
\DeclareFontShape{U}{BOONDOX-calo}{m}{n}{
  <-> s*[1.05] BOONDOX-r-calo}{}
\DeclareFontShape{U}{BOONDOX-calo}{b}{n}{
  <-> s*[1.05] BOONDOX-b-calo}{}
\DeclareMathAlphabet{\mathcalb}{U}{BOONDOX-calo}{m}{n}
\SetMathAlphabet{\mathcalb}{bold}{U}{BOONDOX-calo}{b}{n}
\DeclareMathAlphabet{\mathbcalb}{U}{BOONDOX-calo}{b}{n}

\renewcommand{\paragraph}[1]{{\noindent \textbf{#1.}}}

\newcommand{\namelong}{\textsc{Progressive Inference-Time Annealing}\xspace}
\newcommand{\nameshort}{\textsc{PITA}\xspace}

%% file: macros.tex
\let\originalleft\left
\let\originalright\right
\renewcommand{\left}{\mathopen{}\mathclose\bgroup\originalleft}
\renewcommand{\right}{\aftergroup\egroup\originalright}

\global\long\def\K{\mathcal{K}}

\global\long\def\norm#1{\left\lVert #1\right\rVert }
\global\long\def\inner#1#2{\left\langle #1, #2\right\rangle}

\definecolor{antiquefuchsia}{rgb}{0.57, 0.36, 0.51}
\definecolor{amethyst}{rgb}{0.6, 0.4, 0.8}

\newcommand{\deriv}[2]{\frac{\partial #1}{\partial #2}}

\newcommand{\mean}{\mathbb{E}}
\newcommand{\var}{{\rm I\kern-.3em D}}

\newcommand{\cond}{\,|\,}
\newcommand{\Normal}{\mathcal{N}}

\newcommand{\eps}{\varepsilon}

\newtheorem*{theorem*}{Theorem}
\newtheorem*{proposition*}{Proposition}
\newtheorem*{example*}{Example}
\DeclareMathSymbol{\shortminus}{\mathbin}{AMSa}{"39}
\usepackage{tcolorbox}

\usepackage[framemethod=TikZ]{mdframed}
\mdfdefinestyle{mybox}{%
    linecolor=white,
    outerlinewidth=1pt,
    roundcorner=2pt,
    innertopmargin=10pt,
    innerbottommargin=0.5pt,
    innerrightmargin=10pt,
    innerleftmargin=10pt,
    backgroundcolor=black!3!white}

\tcbuselibrary{theorems}

\newtcbtheorem[number within=section]{mytheo}{Statement}%
{colback=green!5,colframe=green!35!black,fonttitle=\bfseries}{th}

\def\gE{{\mathcal{E}}}
\def\gW{{\mathcal{W}}}
\def\sT{{\mathbb{T}}}

\makeatletter
\let\save@mathaccent\mathaccent
\newcommand*\if@single[3]{%
  \setbox0\hbox{${\mathaccent"0362{#1}}^H$}%
  \setbox2\hbox{${\mathaccent"0362{\kern0pt#1}}^H$}%
  \ifdim\ht0=\ht2 #3\else #2\fi
  }
\newcommand*\rel@kern[1]{\kern#1\dimexpr\macc@kerna}
\newcommand*\widebar[1]{\@ifnextchar^{{\wide@bar{#1}{0}}}{\wide@bar{#1}{1}}}
\newcommand*\wide@bar[2]{\if@single{#1}{\wide@bar@{#1}{#2}{1}}{\wide@bar@{#1}{#2}{2}}}
\newcommand*\wide@bar@[3]{%
  \begingroup
  \def\mathaccent##1##2{%
    \let\mathaccent\save@mathaccent
    \if#32 \let\macc@nucleus\first@char \fi
    \setbox\z@\hbox{$\macc@style{\macc@nucleus}_{}$}%
    \setbox\tw@\hbox{$\macc@style{\macc@nucleus}{}_{}$}%
    \dimen@\wd\tw@
    \advance\dimen@-\wd\z@
    \divide\dimen@ 3
    \@tempdima\wd\tw@
    \advance\@tempdima-\scriptspace
    \divide\@tempdima 10
    \advance\dimen@-\@tempdima
    \ifdim\dimen@>\z@ \dimen@0pt\fi
    \rel@kern{0.6}\kern-\dimen@
    \if#31
      \overline{\rel@kern{-0.6}\kern\dimen@\macc@nucleus\rel@kern{0.4}\kern\dimen@}%
      \advance\dimen@0.4\dimexpr\macc@kerna
      \let\final@kern#2%
      \ifdim\dimen@<\z@ \let\final@kern1\fi
      \if\final@kern1 \kern-\dimen@\fi
    \else
      \overline{\rel@kern{-0.6}\kern\dimen@#1}%
    \fi
  }%
  \macc@depth\@ne
  \let\math@bgroup\@empty \let\math@egroup\macc@set@skewchar
  \mathsurround\z@ \frozen@everymath{\mathgroup\macc@group\relax}%
  \macc@set@skewchar\relax
  \let\mathaccentV\macc@nested@a
  \if#31
    \macc@nested@a\relax111{#1}%
  \else
    \def\gobble@till@marker##1\endmarker{}%
    \futurelet\first@char\gobble@till@marker#1\endmarker
    \ifcat\noexpand\first@char A\else
      \def\first@char{}%
    \fi
    \macc@nested@a\relax111{\first@char}%
  \fi
  \endgroup
}
\makeatother

\newcommand{\xhdr}[1]{{\noindent\bfseries #1}.}
\newcommand{\cut}[1]{}

%% file: text/intro.tex
\section{Introduction}
\label{sec:intro}

\looseness=-1
The problem of sampling from an unnormalized target probability distribution arises in numerous areas of natural sciences, including computational biology, chemistry, physics, and materials science ~\citep{frenkel2023understanding,liu2001monte,ohno2018computational,stoltz2010free,noe2019boltzmann}. In many of these high-impact scientific settings, the problem's complexity stems from operating in molecular systems where the unnormalized target (Boltzmann) distribution at a low temperature of interest is governed by a highly non-convex and non-smooth energy function, under which there is limited to no available data \citep{henin2022enhanced}. As a result, the sampling problem necessitates solving an equally hard exploration problem: finding the modes---in correct proportion---of the target distribution.

\looseness=-1
To address the general sampling problem, extensive research has been dedicated to Markov chain Monte Carlo methods (MCMC), Sequential Monte Carlo (SMC), and, particularly in physical systems, Molecular Dynamics (MD)~\citep{leimkuhler2015molecular}. To enhance scalability, Monte Carlo approaches often employ an interpolating sequence of probability distributions that transitions from an easily sampled reference distribution to the desired target distribution via annealing/tempering strategies. This powerful concept underlies methods such as parallel tempering \citep{swendsen1986replica}, Annealed Importance Sampling \citep{jarzynski1997nonequilibrium,neal2001annealed}, and SMC samplers \citep{Del-Moral:2006}. MD, conversely, involves integrating equations of motion using finely discretized time steps. Despite their effectiveness, both classes of methods possess inherent limitations that complicate their application to practical systems of interest: annealing modifies the masses of distribution modes depending on their widths (a phenomenon often referred to as mass teleportation \citep{woodard2009sufficient}), while MD requires computationally expensive time discretization on the order of femtoseconds to simulate millisecond-scale phenomena.

\looseness=-1
Diffusion-based samplers are an alternative and emergent class of sampling techniques
\citep{zhangyongxinchen2021path,vargasdenoising,akhound2024iterated,berner2022optimal,blessing2024beyond,havens2025adjoint} exploiting modern developments in generative modeling. They sample complex multi-modal distributions by leveraging a prescribed interpolating probability path. However, instead of relying on annealing, these samplers utilize a noising mechanism which theoretically enjoys favorable mixing properties compared to annealing~\citep{chen2023sampling}.%

Diffusion-based samplers, despite their appeal, have not yet proven effective for even small molecular systems in Cartesian coordinates. This is primarily because of the absence of training data to accurately approximate the logarithmic gradients of the marginal densities, i.e. the Stein scores -- a challenge distinct from generative modeling settings. Additionally, standard training objectives, such as reverse Kullback--Leibler, are mode dropping and often yield too high-variance score estimates for stable training~\citep{blessing2024beyond}. Crucially, current diffusion-based samplers require too many energy function evaluations for training. Indeed, when normalized by the number of energy evaluations, carefully tuned MCMC methods with parallel tempering are empirically competitive with, if not superior to, state-of-the-art diffusion-based samplers~\citep{he2025no}.

\begin{figure}[t]
    \centering
    \includegraphics[width=0.99\linewidth]{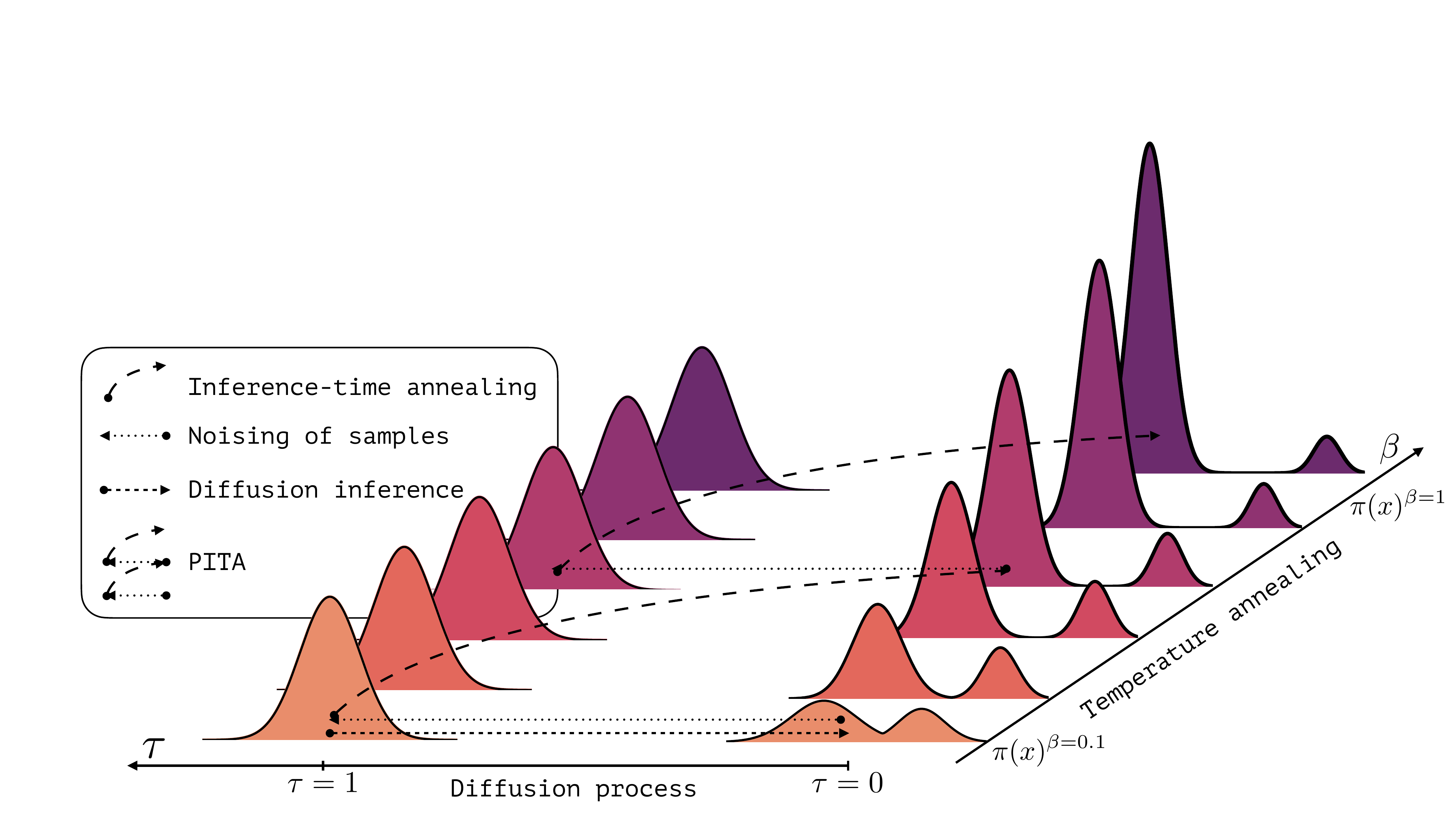}
    \vspace{-5pt}
    \caption{\small Illustration of the proposed \nameshort framework combining two complementary processes: temperature annealing of the target Boltzmann density and the diffusion process applied to the collected samples. Annealed inference allows for decreasing the temperature (increasing $\beta$) of a trained diffusion model, thus generating samples from the annealed target. These samples can be reused for training a lower-temperature diffusion model. }
    \label{fig:pita_framework}
\end{figure}

\looseness=-1
\xhdr{Main Contributions}
In this paper, we introduce \namelong (\nameshort), a novel framework for training diffusion models to sample from Boltzmann distributions. \nameshort leverages two complementary interpolation techniques to significantly enhance training scalability: temperature annealing (increasing the system's temperature) and interpolation along a conventional diffusion model's probability path. This combination is motivated by a learning framework designed to exploit their distinct advantages: temperature annealing mixes modes by lowering high-energy barriers, while diffusion paths avoid mass teleportation.

\looseness=-1
Annealing the target distribution transforms the challenging sampling problem into an easier one by removing high-energy barriers and flattening it. This crucial step enables the cheap collection of an initial high-temperature dataset via classical MCMC, which in turn facilitates the efficient training of an initial diffusion model. Subsequently, we define a novel Feynman-Kac PDE that, when combined with SMC-based resampling, allows us to simulate the trained diffusion model's inference process to produce asymptotically unbiased samples at a lower temperature. This effectively allows us to train the next diffusion model, enabling the progressive and stable training of a sequence of diffusion models until the target distribution is reached, as illustrated in \cref{fig:pita_framework}.

\looseness=-1
We test the empirical performance of \nameshort on standard $N$-body particle systems and short peptides in Alanine Dipeptide and tripeptides. Empirically \nameshort not only achieves state-of-the-art performance in all these benchmarks but is the first diffusion-based sampler that scales to our considered peptides in their native Cartesian coordinates.
More importantly, we demonstrate that progressing down our designed ladder of diffusion models leads to significantly lower energy evaluations compared to MD, which is a step towards realizing the promise of amortized samplers for accelerating equilibrium sampling.

%% file: text/background.tex
\section{Background}
\label{sec:background}

\subsection{Diffusion models}
\label{sec:back_diffusion}
\looseness=-1
A diffusion process defines an interpolating path between an easy-to-sample reference density, such as a multi-variate Normal, and a desired target distribution $\pi(x)$. When samples from the target distribution are available, it is possible to generate samples from intermediate marginals of the diffusion process $p_\tau(x)$ through the following Gaussian convolution:
\begin{align}
    p_\tau(x) = \left(\pi * \Normal(\alpha_\tau y,\sigma_\tau^2 \one)\right)(x) = \int dy\; \Normal(x\cond\alpha_\tau y,\sigma_\tau^2 \one)\pi(y). \label{eq:diffusion_conv}
\end{align}
\looseness=-1
As a result, this means that samples from $p_\tau(x)$ can be generated as $x_\tau = \alpha_\tau y + \sigma_\tau\varepsilon$, where $y \sim \pi(y)$ and $\varepsilon \sim \Normal(0,\one)$. Selecting specific schedules for $\alpha_\tau$ and $\sigma_\tau$ one can ensure the following boundary conditions. For $\tau=0$, $\alpha_\tau = 1,\sigma_\tau = 0$ and $p_{\tau=0}(x) = \pi(x)$, i.e. the marginal matches the target distribution. For $\tau=1$, $\alpha_\tau = 0,\sigma_\tau = 1$ and $p_{\tau=1}(x) = \Normal(0,\one)$, i.e. the marginal matches the standard multivariate normal distribution.

\looseness=-1
Importantly, despite the simplicity of sample generation, the evaluation of density $p_\tau(x)$ is not straightforward, and one has to use deep learning models to approximate either scores $\nabla \log p_\tau(x)$ or marginal densities $p_\tau(x)$. Furthermore, the model density $p_\tau(x)$ or scores $\nabla \log p_\tau(x)$ can be used to generate new samples from $\pi(x)$ using the reverse-time SDE. In particular, the marginals introduced in \cref{eq:diffusion_conv} describe the marginal densities of the forward-time (Ornstein–Uhlenbeck) SDE 
\begin{align}
    dx_\tau = \underbrace{\deriv{\log\alpha_\tau}{\tau}x_\tau}_{\coloneqq a_{1-\tau}x_\tau} d\tau + \underbrace{\sqrt{2\sigma^2_\tau \deriv{}{\tau}\log \frac{\sigma_\tau}{\alpha_\tau}}}_{\coloneqq \zeta_{1-\tau}} d\widebar{W}_\tau\,, \;\; x_{\tau=0} \sim \pi(x)\,,
    \label{eq:noising_sde}
\end{align}
where $\widebar{W}_\tau$ is the standard Wiener process, and marginals follow the Fokker-Planck PDE. After inverting the time variable in this PDE, i.e. $t = 1-\tau$, the time-evolution of marginals $p_t(x)$ is
\begin{align}
    \deriv{p_{t}(x)}{t} = \inner{\nabla}{p_{t}(x)\left(a_t x\right)} - \frac{\zeta_t^2}{2}\Delta p_{t}(x) = -\inner{\nabla}{p_{t}(x)(-a_tx + \frac{\zeta_t^2}{2}\nabla \log p_{t}(x))}\,,\label{eq:ref_FP}
\end{align}
which shows that one can sample from the marginals $\{p_t(x)\}_{t\in[0,1]}$ via simulating the following SDE
\begin{align}
    dx_t = \left(-a_t x + \frac{\zeta_t^2}{2}(1+\xi_t)\nabla \log p_{t}(x_t)\right)dt + \zeta_t \sqrt{\xi_t} dW_t\,.
    \label{eq:reverse_SDE}
\end{align}
\looseness=-1
While the marginals are correct for any $\xi_t > 0$, there are two important special cases: for $\xi_t \equiv 1$, the equation becomes the reverse-SDE with the same path-measure as~\cref{eq:noising_sde}, and, for $\xi_t \equiv 0$, the SDE becomes an ODE. In practice, this SDE is simulated using either the density model $\exp(-U_t(x;\eta)) \propto p_t(x)$ \citep{du2023reduce} or the score model $s_{t}(x;\theta) \approx \nabla \log p_{t}(x)$ \citep{song2020score}. 

\subsection{Annealing}
\label{sec:back_annealing}

Annealing defines a family of ``simpler'' problems when we have access to the unnormalized density by interpolating or scaling the target log-density (negative energy). Formally speaking, given the unnormalized density $\pi(x)$, the annealed density is defined as
\begin{align}
    \pi^\beta(x) \coloneqq \frac{\pi(x)^\beta}{Z_\beta}\,,\;\; Z_\beta = \int dx\; \pi(x)^\beta\,,
\end{align}
where $\beta$ is the inverse temperature parameter (i.e., $\beta = 1/T$) controlling the smoothness of the target density. Thus, for high temperature $T > 1$ (hence, $\beta < 1$), the target density becomes smoother and easier to explore via MCMC algorithms. Importantly, getting the unnormalized density for $\pi^\beta(x)$ can be simply achieved by raising the unnormalized density $\pi(x)$ to the power $\beta$.

\subsection{Feynman-Kac Formula}
\label{sec:back_feynmankac}

The Feynman-Kac Partial Differential Equation (PDE) is a generalization of the Fokker-Planck PDE and is defined as follows
\begin{align}
    \deriv{p_{t}(x)}{t} = -\inner{\nabla}{p_{t}(x)v_t(x)} + \frac{\zeta_t^2}{2}\Delta p_{t}(x) + p_t(x)\left(g_t(x) - \mean_{p_t(x)}g_t(x)\right)\,,
\end{align}
where the first term corresponds to the probability mass transport along the vector field $v_t(x)$, the second term corresponds to the stochastic moves of samples according to the Wiener process $W_t$, and the last term is responsible for reweighting the samples according to a coordinate dependent weighting function $g_t(x)$. For any test-function $\varphi(x)$, the Feynman-Kac formula relates its expected value to the expectation over the SDE trajectories $x_t$, i.e.
\begin{align}
    \mean_{p_T(x)}[ \varphi(x) ] = \frac{1}{Z_T} \mean\left[ e^{\int_0^T dt\; g_t(x_t)} \varphi(x_T) \right]\,,\;\text{where}\; dx_t = v_t(x_t) dt + \zeta_t dW_t\,,\; x_0 \sim p_0(x)\,,
    \label{eq:fk_formula}
\end{align}
\looseness=-1
and $Z_T$ is a normalization constant independent of $x$. In practice, the exponential term is computed as a ``weight'' $w_t$ of the corresponding sample $x_t$ and can be integrated in parallel with the simulation,
\begin{align}
    dx_t =~& v_t(x_t)dt + \zeta_t dW_t\,,\;\; d\log w_t = g_t(x_t)dt\,,\;\text{ initialized as }\; x_0 \sim p_0(x)\,,\;\; \log w_0 = 0\,.
    \label{eq:weighted_SDE}
\end{align}
Finally, one can estimate the normalization constant $Z_T$ by considering $\varphi(x) \equiv 1$ in \cref{eq:fk_formula} and get the biased but consistent Self-Normalized Importance Sampling (SNIS) estimator \citep{liu2001monte}, i.e.
\begin{align}
    \frac{1}{Z_T} \mean\left[ e^{\int_0^T dt\; g_t(x_t)} \varphi(x_T) \right] = \frac{\mean e^{\int_0^T dt\; g_t(x_t)} \varphi(x_T)}{\mean e^{\int_0^T dt\; g_t(x_t)}} \approx \sum_{i=1}^n \frac{w_T^i}{\sum_{j=1}^n w_T^j}\varphi(x_T^i)\,,
    \label{eq:snis}
\end{align}
where $(x_T^i, w_T^i)$ are the samples from the simulation of the SDE in \cref{eq:weighted_SDE}.

%% file: text/method.tex
\section{Progressive Inference-Time Annealing}
\label{sec:method}
\looseness=-1
In this section, we combine diffusion and annealing processes into an efficient learning algorithm for sampling from the target density $\pi(x)$. To design this method, we build on the fact that diffusion and annealing are complementary ways to simplify or ``smoothen'' the target distribution (see \cref{fig:pita_framework}). Namely, for the high-temperature version of the target distribution $\pi^{\beta_i}(x)$, we assume having samples from $\pi^{\beta_i}(x)$ and learn the density model of the marginals defined by the diffusion process (see \cref{sec:training}). For instance, this can be done by running MCMC chains that face little challenge mixing in high temperatures. For the given density model of the diffusion process, we perform annealing of all the marginals and generate samples from the lower temperature target $\pi^{\beta_{i+1}}(x)\,,\beta_{i+1} < \beta_{i}$ (see \cref{sec:inference}). We detail every step of our method in the following subsections.

\subsection{Inference-Time Annealing}
\label{sec:inference}
In this section, we discuss the inference-time annealing process, which allows us to modify the trained diffusion model to generate samples from the lower temperature target density. Namely, for a diffusion process with marginals $p_t(x)$ and the end-point $p_{t=1}(x)=\pi^{\beta_i}(x)$, we assume having two models: a score model $s_{t}(x;\theta) \approx \nabla \log p_t(x)$ and an energy-based model $U_{t}(x;\eta) \approx -\log p_t(x) + \text{const}$ with parameters $\theta$ and $\eta$ respectively. Given the score and the energy models trained to sample from a higher temperature density $\pi^{\beta_i}(x)$, we define a new sequence of marginals that correspond to the Boltzmann density of the energy model but with a lower temperature
\begin{align}
    q_{t}(x) \propto \exp\left(-\beta_{i+1}/\beta_iU_{t}(x;\eta)\right)\,,\; q_{t=1}(x) \propto \exp\left(-U_{t=1}(x;\eta)\right)^{\beta_{i+1}/\beta_i} \approx (\pi(x)^{\beta_i})^{\beta_{i+1}/\beta_i}\,.
\end{align}
The following proposition derives the Feynman-Kac PDE that describes the time-evolution of the marginals $q_t(x)$ and allows for importance sampling via the Feynman-Kac formula.
\begin{mdframed}[style=mybox]
\begin{restatable}{proposition}{annealing}
\label{prop:annealing}
{\textup{[Inference-time Annealing]}}
    Annealed density of the energy-based model $q_t(x) \propto \exp\left(-\gamma U_{t}(x;\eta)\right)$ matches the marginal densities of the following SDE
    \begin{align}
        dx_t =~& \left(-a_t x_t+\frac{\zeta_t^2}{2}\left(s_{t}(x_t) - \gamma\xi_t\nabla U_{t}(x_t;\eta)\right)\right)dt + \zeta_t\sqrt{\xi_t} dW_t\,,\;\; x_0 \sim q_{t=0}(x) \label{eq:states_sde}\\
        d\log w_t =~&  \left[\frac{\zeta_t^2}{2}\inner{\nabla}{s_{t}(x_t)} -\gamma\inner{\nabla U_{t}(x_t;\eta)}{-a_t x_t + \frac{\zeta_t^2}{2}s_{t}(x_t)} - \gamma\deriv{U_{t}(x_t;\eta)}{t}\right]dt\,,\label{eq:weights_sde}
    \end{align}
    where $s_{t}(x)$ is any vector field, $a_t,\zeta_t,\xi_t$ are analogous to parameters from \cref{eq:reverse_SDE}, and the sample weights $w_t$ correspond to the SNIS estimator of the Feynman-Kac formula in \cref{eq:snis}.
\end{restatable}
\end{mdframed}
See \cref{app:annealing} for the proof. Intuitively, this result defines an importance sampling scheme, where \cref{eq:states_sde} generates samples from the proposal distribution and \cref{eq:weights_sde} integrates the density ratio between the sampled proposal and the target density $q_t(x)$. Different choices of the vector field $s_t(x)$ and the noise schedule $\xi_t$ yield different proposal distributions. Theoretically, one can choose different parameters $a_t,\zeta_t$ as well, but below we argue for setting them according to \cref{eq:noising_sde}.

\looseness=-1
The dynamics in \cref{prop:annealing} are not unique, and there exists a continuous family of PDEs that follow the marginals $q_t(x) \propto \exp(-\gamma U_t(x;\eta))$. We motivate this dynamics as minimizing the variance of the weights for the case when there is no annealing ($\gamma = 1$). Indeed, if the trained EBMs and score models approximate the diffusion process perfectly, then, for $\gamma = 1$, the weights become constant, so SNIS equally weights all the samples; thus, eliminating the need to resample at all. We formalize this result in the following proposition (see \cref{app:annealing} for the proof).
\begin{mdframed}[style=mybox]
\begin{restatable}{proposition}{annealingvar}
\label{prop:annealingvar}
{\textup{[Convergence to Diffusion]}}
    For $\gamma=1$ and perfect models $s_{t}(x) = -\nabla U_{t}(x;\eta) = \nabla\log p_{t}(x)$, the variance of the weights in \cref{prop:annealing} becomes zero.
\end{restatable}
\end{mdframed}

\looseness=-1
In the case of unbounded support of the target distribution, e.g. $\texttt{supp}(\pi(x)) = \mathbb{R}^d$, increasing the temperature might cause numerical instabilities. Indeed, $\pi(x)^{\beta=0} \propto \texttt{Uniform}(\R^d)$ is not normalizable. To avoid this potential issue, in \cref{app:averaging}, we consider geometric averaging between some simple prior and the target densities, e.g. $\Normal(0,\one)^{(1-\beta)}\pi(x)^\beta$. 

\looseness=-1
Integrating the dynamics from \cref{prop:annealing,prop:averaging} we generate a set of weighted samples $\{(x^k_{t=1},w^k_{t=1})\}_{k=1}^K$ that converge to the samples from $q_{t=1}(x)$ when $K \to \infty$. In practice, this density is defined as the Boltzmann distribution of the corresponding energy model, i.e. $q_{t=1}(x) \propto \exp(-\beta_{i+1}/\beta_i\cdot U_{t=1}(x;\eta))$, which approximates $\pi^{\beta_{i+1}}(x)$, but does not necessarily match it exactly. We discuss several possible ways to bridge this gap between the density model and the target density in \cref{app:rendezvous}.

\subsection{Training using \nameshort}
\label{sec:training}
The proposed algorithm consists of interleaving the inference-time annealing (described in the previous two subsections) and model training on the newly generated data from the annealed target distribution, which we describe here. Throughout this stage we assume availability of samples from $\pi^{\beta_{i+1}}$, which were previously generated at the sampling stage\footnote{For the very first iteration of our algorithm, we assume that there exist such $\beta$ that samples from $\pi^\beta$ can be simply collected by conventional Monte Carlo algorithms.}.

\looseness=-1
For the target distribution $\pi^{\beta_{i+1}}(x)$, we define the diffusion process with the marginals $p_t(x)$ obtained as a convolution of the samples from the target $x\sim\pi^{\beta_{i+1}}(x)$ with the Gaussian $\Normal(\alpha_\tau x,\sigma_\tau^2\one)$. To learn the score function $s_{t}(x;\theta) \approx \nabla \log p_{t}(x)$, we follow the standard practice and parameterize the denoising model $D_t(x_t;\theta)=\sigma^2_t s_t(x_t;\theta)+x_t$, which we learn via the denoising score matching (DSM) objective \citep{ho2020denoising}, i.e.
\begin{align}
    \text{Denoising Score Matching}(\theta) =~& \mean_{t,x_t,x} \lambda(t)\norm{x - D_{t}(x_t;\theta)}^2\,,
    \label{eq:score_matching}
\end{align}
where the expectation is taken w.r.t. samples from the annealed target $x\sim \pi^{\beta + \Delta\beta}(x)$, noised samples $x_t \sim \Normal(x_t\cond \alpha_{1-t}x, \sigma^2_{1-t})$, and time parameter sampled with $\log (1-t)\sim \mathcal{N}(P_{mean}, P_{std})$ largely following \citet{karras2022elucidating}.

\begin{algorithm}[t]
\caption{Training for single temperature $1/\beta_{i+1}$}
\label{alg:training}
\begin{algorithmic}
\REQUIRE samples $x$ from $\pi^{\beta_{i+1}}$.
\FOR{training iterations}
\STATE sample $\ln (\sigma_{1-t}) \sim \Normal(P_{mean}, P_{std}^2)$
\STATE add noise $x_t \sim \Normal(x_t\cond \alpha_{1-t} x, \sigma^2_{1-t}\one)$
\STATE $\text{Denoising Score Matching}(\theta) = \nabla_\theta\;\mean_{t,x_t,x} \lambda(t)\norm{x - D_{t}(x_t;\theta)}^2$
\STATE $\text{Target Score Matching}(\theta) = \nabla_\theta\;\mean_{t,x_t,x} \left [\norm{\sigma_t^2\nabla_{x} \log \pi(x) + x - D_{t}(x_t;\theta)}^2 \cdot \mathbb{I}(t \geq t_{\text{thresh}}) \right ]$
\STATE $\text{EBM Distillation}(\eta) = \nabla_\eta\;\mean_{t,x_t,x} \lambda(t) \norm{\sigma_t^2(-\nabla_{x_t} U_{t}(x_t;\eta)) + x_t - D_{t}(x_t;\theta)}^2$
\STATE $\text{Energy Pinning}(\eta) = \nabla_\eta\;\mean_{x} \norm{(-U_{t=1}(x;\eta)) - \beta_{i+1}\log \pi(x)}^2$
\STATE $\theta \gets \texttt{FirstOrderOptimizer}(\theta, \text{Score Matching}(\theta), \text{Target Score}(\theta))$
\STATE $\eta \gets \texttt{FirstOrderOptimizer}(\eta, \text{Energy Matching}(\eta), \text{Energy Pinning}(\eta))$
\ENDFOR
\RETURN trained parameters $\theta^*,\eta^*$
\end{algorithmic}
\end{algorithm}
\looseness=-1
However, the DSM objective is not sufficient for training a good score model close to the target distribution ($\tau = 1-t = 0$) due to the high variance of the estimator. Indeed, for $\tau = 1-t = 0$, it has no information about the target distribution. Target Score Matching \citep{de2024target} overcomes this issue by explicitly incorporating the score of the target unnormalized density into the objective, which is as follows
\begin{align}
    \text{Target Score Matching}(\theta) = \mean_{t,x_t,x} \left [\norm{\sigma_t^2\nabla_{x} \log \pi(x) + x - D_{t}(x_t;\theta)}^2 \cdot \mathbb{I}(t \geq t_{\text{thresh}}) \right ]
\end{align}
where the expectation is taken w.r.t.\ the same random variables as in \cref{eq:score_matching}, but the time variable is restricted to the interval $[t_{\text{thresh}},1]$ because the variance of the objective estimator grows with the noise scale~\citep{de2024target}. Notably, for larger noise scales, the Denoising Score Matching objective results in a stable training dynamics; thus, these objectives complement each other and result in a stable training dynamics across the entire time interval.

\looseness=-1
To train the energy model $U_{t}(x;\eta)$, which plays the central role in the inference-time annealing (see \cref{sec:inference}), we follow \cite{thornton2025composition} and distill the learned score model to the parametric energy model via the following regression loss (w.r.t.\ $\eta$), i.e.
\begin{align}
    \text{EBM Distillation}(\eta) = \mean_{t,x_t,x} \lambda(t)\norm{\sigma_t^2(-\nabla_{x_t} U_{t}(x_t;\eta)) + x_t - D_{t}(x_t;\theta)}^2\,,
    \label{eq:energy_matching}
\end{align}
where, the expectation is taken w.r.t.\ the same random variables as in \cref{eq:score_matching}. Note that, in contrast to the denoising score matching loss in \cref{eq:score_matching}, the ``target'' in \cref{eq:energy_matching} does not depend on $x$, which means that its variance for the same $x_t$ is zero, stabilizing the training of the energy based model $U_{t}(x;\eta)$.

\looseness=-1
Finally, to use all the supervision signal available in the problem, we use the target unnormalized density $\pi(x)^{\beta_{i+1}}$ as the regression target for the end-point energy-based model $U_{t=1}(x;\eta)$, and introduce the following loss
\begin{align}
    \text{Energy Pinning}(\eta) = \mean_{\pi^{\beta_{i+1}}(x)} \norm{(-U_{t=1}(x;\eta)) - \beta_{i+1}\log \pi(x)}^2\,.
    \label{eq:target_matching}
\end{align}
Notably, this loss allows for fixing the gauge present due to the shift invariance of the energy-based model ($\nabla_x(U_t(x;\eta)) = \nabla_x(U_t(x;\eta) + \text{const})$). In practice, we observe that this loss significantly stabilizes the training and improves the final performance.
\looseness=-1
We present the pseudo-code for the training loop in \cref{alg:training}, where we simultaneously optimize all the introduced loss functions to train a diffusion model at temperature $1/\beta_{i+1}$. In practice, we find that sequential training of models demonstrates the best performance. Furthermore, we initialize the model for the next temperature $1/\beta_{i+1}$ with the parameters of the trained model for the temperature $1/\beta_i$.

%% file: text/related_work.tex
\section{Related work}
\label{sec:related_work}

\looseness=-1
\xhdr{Diffusion-based Sampling}
A variety of amortized samplers that use properties of diffusion models have recently been proposed in the literature. Simulation-based approaches 
that also exploit the fast mode-mixing of diffusion models include~\citet{berner2022optimal,vargasdenoising,zhangyongxinchen2021path,richter2023improved,vargas2024transport}, which exploit diffusion processes for fast mode mixing. Conversely, simulation-free methods like iDEM~\citep{akhound2024iterated}, SB with F\"ollmer drift~\citep{huang2021schrodinger}, and TSM~\citep{de2024target} offer more scalable approaches to learning the score but suffer from inefficient and high variance score estimates far from the data. Finally, new diffusion bridges have also risen to prominence with underdamped dynamics~\citep{blessing2024beyond}, known modes~\citep{noble2025learned}, and bridges with SMC~\citep{chen2024sequential}.

\xhdr{Inference-time Resampling} The inference-time annealing scheme proposed in \cref{prop:annealing} connects several recently proposed methods. Namely, for $\xi_t\equiv 0$, it closely matches the importance sampling of the continuous normalizing flows proposed in \citep{kohler2020equivariant}. Indeed, \cref{eq:states_sde} becomes a probability flow ODE, and \cref{eq:weights_sde} becomes an integration of the log-density-ratio, where the target density can be defined either as a linear interpolation of log-densities or only in the final point as the target density. Furthermore, \cref{prop:annealing} is an application of the Feynman-Kac formula to annealing and non-equilibrium dynamics simultaneously. Indeed, for $\gamma = 1$, this proposition becomes the result proposed in \citet{vaikuntanathan2008escorted,albergo2024nets}; whereas, for $s_t(x) = -\nabla U_t(x;\eta) = \nabla\log p_t(x)$, this proposition becomes the result from \citep{skreta2025feynman}. In practice, however, these equalities are not satisfied because we use learned models for the vector field $s_t(x) = s_t(x;\theta)$ and the energy-based model $U_t(x;\eta)$. In concurrent work, \citet{rissanen2025progressivetemperingsamplerdiffusion} also propose an annealed sampling scheme using diffusion models. They use a different inference-time annealing procedure which restricts importance re-sampling to the final timestep, whereas in this work, we benefit from \textit{annealed} importance sampling over diffusion time. 

\looseness=-1
\xhdr{Boltzmann Generators} \citet{noe2019boltzmann} proposed training a probabilistic model and resampling the generated samples according to the target Boltzmann density via importance sampling. Various probabilistic models have been used, e.g.\ continuous normalizing flows \citep{chen2018neural} and the flow matching objective \citep{lipman2022flow} also allow for efficient training and resampling under the Boltzmann generators framework \citep{kohler2020equivariant, klein2023equivariant}. Boltzmann Generators can also be combined with Annealed Importance Sampling, which enhances their scalability~\citep{tan2025scalable}.
However, as we demonstrate empirically, the straightforward resampling with a target density of a different temperature~\citep{schopmans2025temperatureannealedboltzmanngenerators} results in high variance of importance weights. Thus, one has to deviate from the Boltzmann Generators framework to perform the inference-time annealing.

%% file: text/experiments.tex
\section{Experiments}
\label{sec:experiments}
We evaluate \nameshort on molecular conformation sampling tasks including a toy Lennard-Jones system of 13 particles (LJ-13) and Alanine peptide systems of varying sizes (Alanine Dipeptide and Tripeptide) in Cartesian coordinate space. Throughout, we assume access to a short MCMC chain run at high temperature. Note that we do not require these chains to be well mixed, but only require them to cover the modes, a much less stringent requirement (See \cref{app:md-mixing}). For metrics, we use sample-based metrics such as 2-Wasserstein distance on Ramachandran coordinates ($\sT$-$\W_2$) and energy distribution ($\cE$-$\W_1$, $\cE$-$\W_2$), to assess mode coverage and precision, respectively. We also compare the KL divergence between the Ramachandran plots of the ground-truth MD samples and the generated samples (RAM-KL), as well as Wasserstein distances on the first two TICA (Time-lagged Independent Component Analysis) coordinates of ground-truth and generated samples. Finally, we report the computational expense of all methods using the total number of energy evaluation calls.

\xhdr{Baselines} 
We compare \nameshort with three different baselines: Temperature Annealed Boltzmann Generators (TA-BG,  \cite{schopmans2025temperatureannealedboltzmanngenerators}), normalizing flow model trained on molecular simulation (MD) data collected at the target temperature (MD-NF), diffusion model trained on molecular simulation (MD) data collected at the target temperature (MD-Diff), and importance sampling using continuous normalizing flows \citep{kohler2020equivariant}. For the LJ-13 dataset, we additionally compare the performance of the model with two state-of-the-art diffusion-based sampling algorithms: namely, iDEM \citep{akhound2024iterated} and adjoint sampling \citep{havens2025adjoint}; however, as none of the current diffusion-based approaches are able to achieve competitive performance on the small protein tasks, we only compare to the TA-BG baseline for those datasets. Additionally, for the ALDP experiment, we compare against two other annealing strategies for the diffusion-only approach: FKC and Score Scaling. FKC refers to the annealing scheme in \cite{skreta2025feynman}, and score scaling is essentially scaling the score with the corresponding temperature factor $\gamma$. Further experimental details of the baselines, as well as additional baselines, are provided in the \cref{app:experiment_details,app:additional_baselines}.

\xhdr{Architecture}
In training \nameshort, we use EGNN \citep{satorras2022enequivariantgraphneural} as the model backbone for LJ-13, and DiT~\citep{peebles2023scalablediffusionmodelstransformers} for Alanine Dipeptide and Alanine Tripeptide. Our training follows a sequential temperature schedule, proceeding from high to low temperatures. After training at a given temperature for a fixed number of epochs, we generate samples at the next lower temperature and continue training at that temperature. For LJ-13, we train a single diffusion model conditioning it on $\beta$ and using the data across all previously seen temperatures. For molecular conformation sampling tasks, we adopt a fine-tuning approach, where at each temperature step, the model is trained only on the newly generated samples corresponding to the current temperature without revisiting earlier ones. For the TA-BG baseline, we train TarFlow \citep{zhai2024normalizingflowscapablegenerative} with adaptations suited to molecular data for all three systems, and for MD-Diff, we use the same DiT architecture we used in \nameshort. We parameterize the energy network using the parameterization in \citet{thornton2025composition} and use the preconditioning ($c_s$, $c_{out}$, $c_{in}$, $c_t$) of \cite{karras2022elucidating} for both energy and score networks. Further training details and hyperparameters are provided in \cref{app:experiment_details}.

\xhdr{Hyperparameters} \cref{prop:annealing} allows for many choices of the vector field $s_t(x)$. In practice, we set it proportional to the score model $s_t(x) \propto s_t(x;\theta)$ and try several scaling coefficients (see \cref{app:additional_ablations}). Finally, one can easily add the time-dependent schedule $\gamma_t$ by adding the extra term to the weights, i.e. $d\log w_t = \text{\cref{eq:weights_sde}} - U_{t}(x_t;\eta)\partial\gamma_t/\partial t dt$, which we study in \cref{app:additional_ablations}. We use the noise schedule from \cite{karras2022elucidating}, where, for all experiments, we set $\sigma_{\mathrm{max}}=80$ and $\rho=7$. For LJ-13 and molecular conformation sampling tasks, we use $\sigma_{\mathrm{min}}=0.05, 0.01$, respectively.

\subsection{Main results} \label{sec:main_results}
\begin{figure}[t!]
\vspace{-10pt}
  \centering
  \begin{subfigure}[t]{0.49\linewidth}
    \centering
    \includegraphics[width=\linewidth]{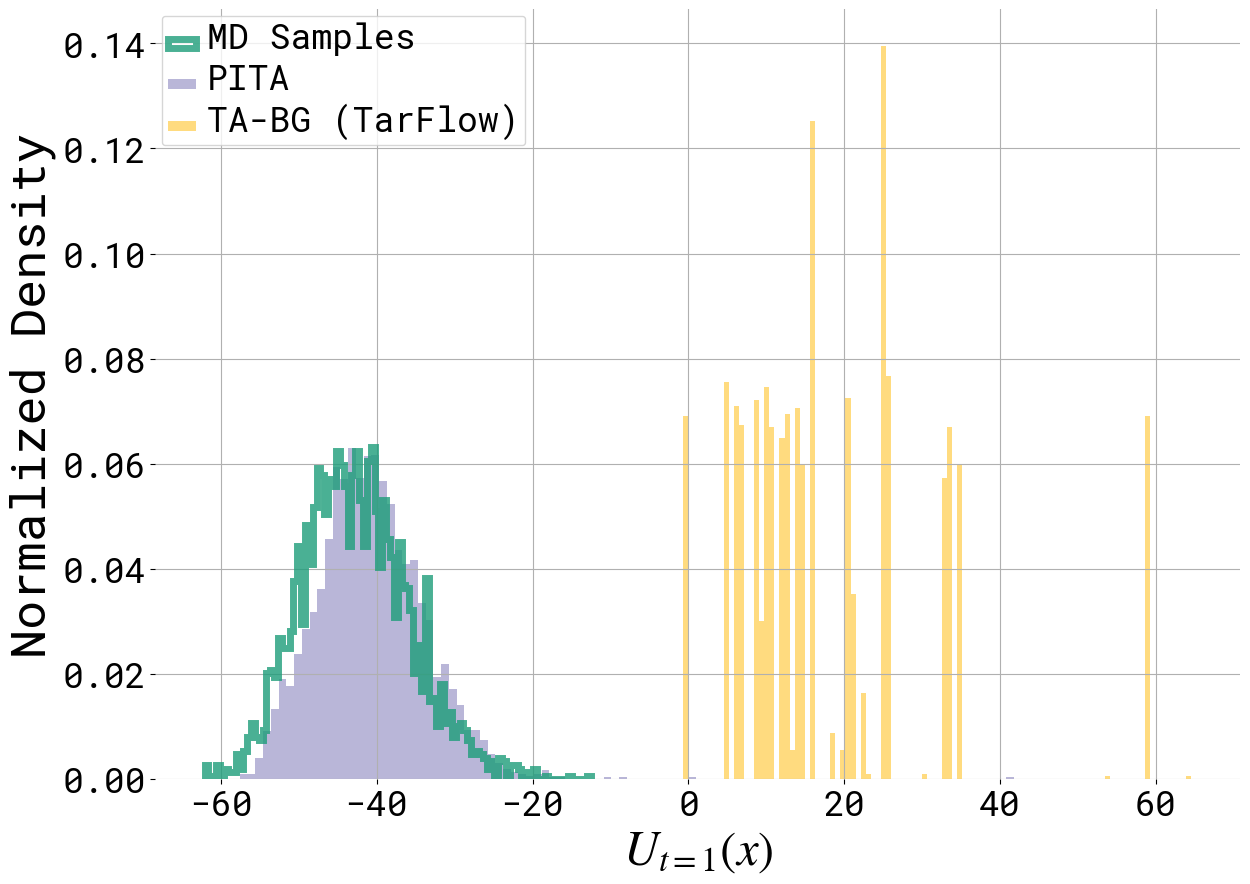}
    \label{fig:lj13_energy}
  \end{subfigure}
  \hfill
  \begin{subfigure}[t]{0.49\linewidth}
    \centering
    \includegraphics[width=\linewidth]{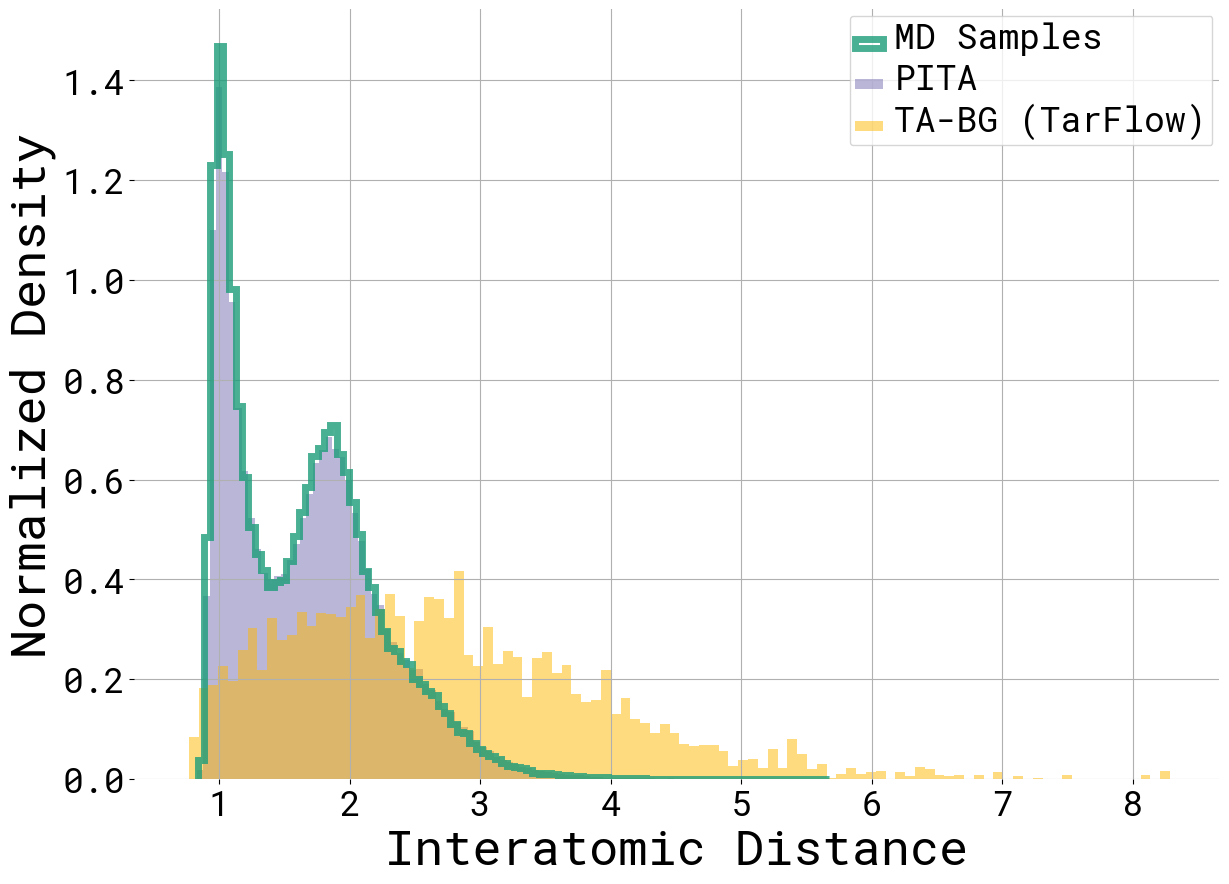}
    \label{fig:lj13_distance}
  \end{subfigure}
\vspace{-10pt}
  \caption{LJ-13 sampling task. We compare the distribution of the interatomic distances and energy of the particles in the MCMC dataset (ground-truth), samples generated using a PITA model, and TA-BG progressively trained from high temperature to sample from the target distribution.}
  \label{fig:lj13_comparison}
\vspace{-10pt}
\end{figure}

\begin{wraptable}{r}{0.6\textwidth}
\vspace{-15pt} 
\centering
\caption{\small LJ-13 sampling task. The starting temperature is $T_L = 4$, annealed to $T_S = 1$.}
\label{tab:lj_results}
\resizebox{0.6\textwidth}{!}{%
\begin{tabular}{@{}lrrr@{}}
\toprule
\textbf{Algorithm} & Distance-$\gW_2$ ↓ & Energy-$\gW_2$ ↓ & Geometric-$\gW_2$ ↓        \\
\midrule
iDEM              & 0.127        & 30.78 ± 24.46  & \textbf{1.61 ± 0.01}            \\
Adjoint Sampling  & -                     & 2.40 ± 1.25   & 1.67 ± 0.01         \\
TA-BG (TarFlow)   & 1.21 ± 0.02           & 61.47 ± 0.12   & 4.16 ± 0.01        \\
\textbf{\nameshort (Ours)} & \textbf{0.04 ± 0.00}  & \textbf{2.26 ± 0.21} & 1.65 ± 0.00       \\
\bottomrule
\end{tabular}
}
\vspace{-10pt}
\end{wraptable}

\textbf{LJ-13.} We first consider a Lennard-Jones (LJ) system of 13 particles to demonstrate the effectiveness of training a sampler at a high temperature ($T_L = 4$), followed by annealing to a lower temperature ($T_S = 1$). As shown in \cref{tab:lj_results}, we compare the performance of \nameshort with existing baselines. A visual comparison to TA-BG is provided in \cref{fig:lj13_comparison}. We evaluate each method using the 2-Wasserstein distance computed over interatomic distance distributions, energy distributions, and 3D geometric coordinates. We omit the Distance-$\gW_2$ metric for Adjoint Sampling, as its results could not be reproduced and no code is available at this time; the reported Energy- and Geometric-$\gW_2$ values are taken directly from the original paper~\citep{havens2025adjoint}. To ensure consistency, we exclude samples with energy above 1000 across all methods; this notably impacts TA-BG, removing approximately 60\% of its samples. Even under this filtering, \nameshort consistently outperforms TA-BG and other baselines trained directly at the target temperature. 

\begin{figure}[htbp]
  \centering
  \includegraphics[width=\linewidth]{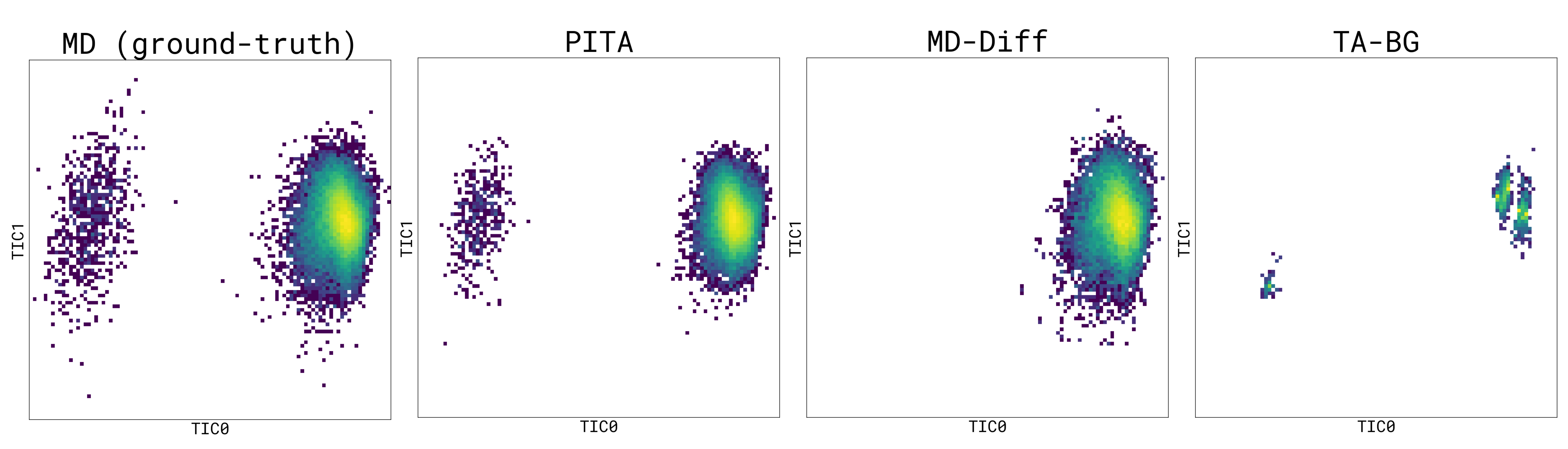}
  \caption{TICA plots for Alanine Dipeptide (ALDP) at 300K obtained from different methods using 30k samples. Each panel shows the free energy landscape along the top two TICA components which capture the dominant slow transitions in the system.}
  \label{fig:tica_aldp}
\end{figure}

\xhdr{Alanine Dipeptide}
We apply \nameshort to the task of sampling Alanine Dipeptide at a target temperature of $T_S = 300$ K, given initial samples at a higher temperature of $T_L = 1200$ K. We use annealing steps of ${1200\ \mathrm{K},\ 755.95\ \mathrm{K},\ 555.52\ \mathrm{K},\ 300\ \mathrm{K}}$. These temperatures correspond to a subset of the temperatures from \cite{schopmans2025temperatureannealedboltzmanngenerators}, as \nameshort does not require as many annealing steps to achieve competitive performance. We also analyze the performance of the model, taking larger annealing steps in \cref{app:additional_ablations}. As shown in \cref{tab:aldp_tab}, \nameshort consistently outperforms both the diffusion-based baseline and TA-BG across all evaluation metrics, achieving a particularly large margin in energy-related metrics. We further present TICA plots of the generated samples at the target temperature in \cref{fig:tica_aldp}. \nameshort successfully recovers the essential slow collective dynamical modes of the system, which baseline methods fail to capture. Additionally, we find that while TA-BG performs reasonably well at earlier stages of training at higher temperatures, its performance deteriorates as temperature decreases. Such a decline is likely due to the increasing difficulty in generating high-quality proposals as the temperature decreases, which is crucial in the importance sampling used for subsequent training stages, as well as the difficulty in importance sampling between large temperature gaps (\cref{app:temp_ess}). Additional details on training dynamics across all temperatures for \nameshort and TA-BG can be found in \cref{app:analysis_temperature}.

\begin{figure}[t!]
  \centering
  \begin{subfigure}[t]{0.49\linewidth}
    \centering
    \includegraphics[width=\linewidth]{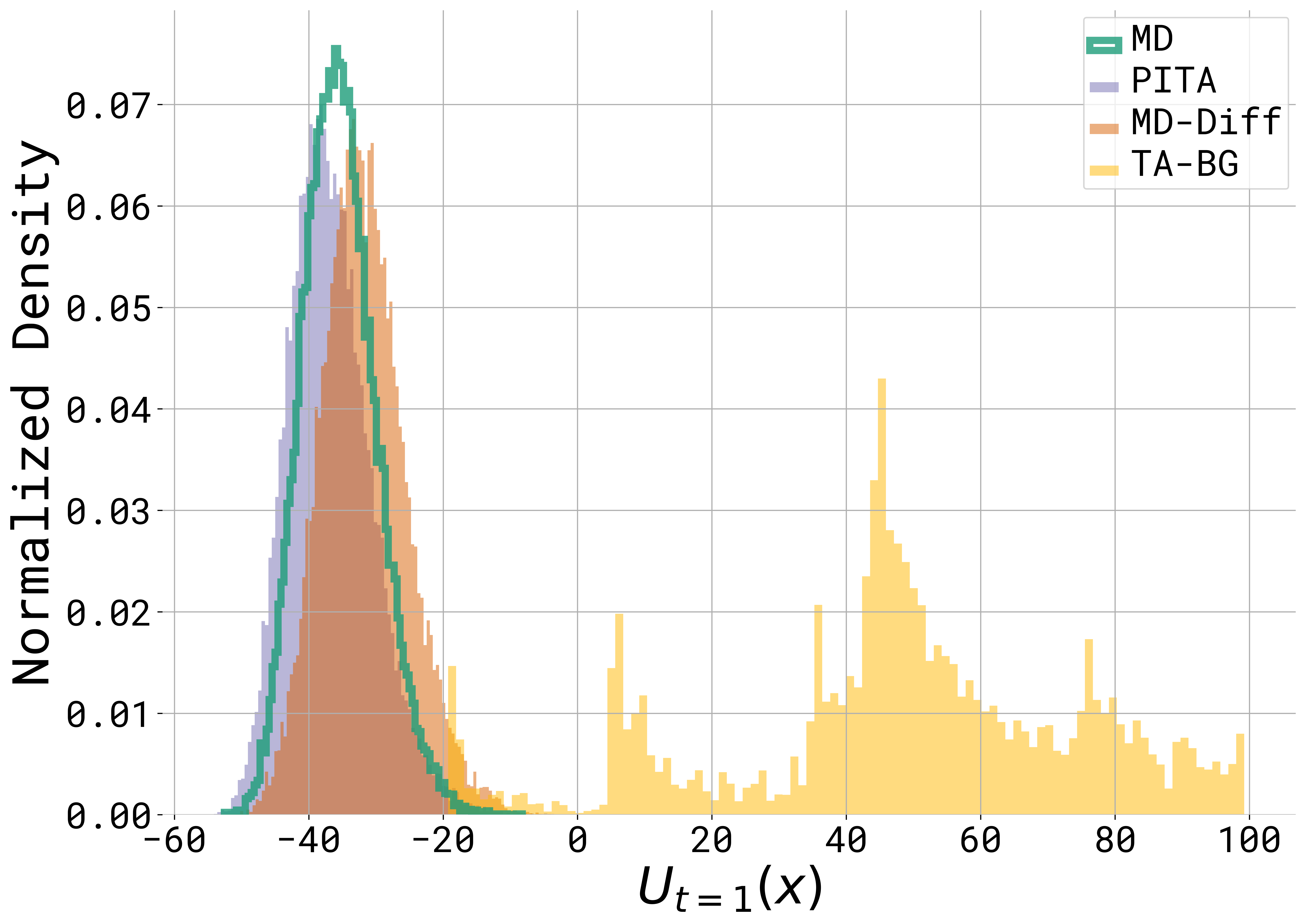}
    \caption{Alanine Dipeptide}
    \label{fig:aldp_energy}
  \end{subfigure}
  \hfill
  \begin{subfigure}[t]{0.49\linewidth}
    \centering
    \includegraphics[width=\linewidth]{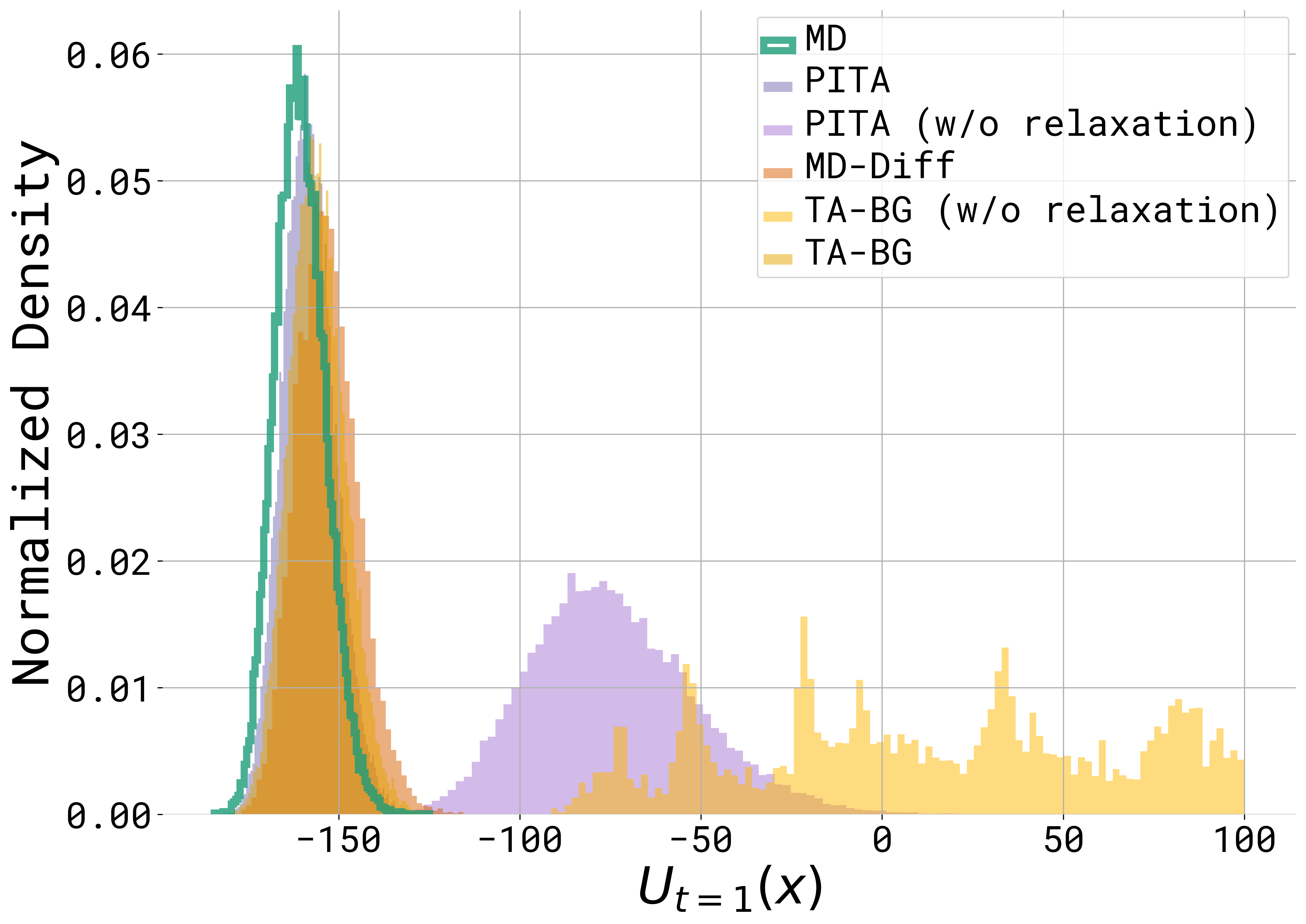}
    \caption{Alanine Tripeptide}
    \label{fig:al3_energy}
  \end{subfigure}
  \caption{Molecular conformation sampling tasks. We compare the energy distribution of the ground-truth MD dataset and the samples generated using different models at $300$K. We use 30k samples for the plots.}
  \label{fig:alp_comparison}
\end{figure}

\begin{table}[t!]
\caption{\small Performance of methods for the ALDP sampling task. The starting temperature is $T_L = 1200$ K, annealed to target $T_S = 300$ K. Metrics are calculated over 10k samples and standard deviations over 3 seeds.}
\resizebox{\textwidth}{!}{%
\begin{tabular}{lrrrrrrr}
\toprule
 & Rama-KL & Tica-$\gW_1$ $\downarrow$ & Tica-$\gW_2$ $\downarrow$ &Energy-$\gW_1$ $\downarrow$ & Energy-$\gW_2$ $\downarrow$ & $\mathbb{T}$-$\mathcal{W}_2$  &  \#Energy Evals \\
\midrule
\nameshort & 4.773 $\pm$ 0.460 & \textbf{0.112 $\pm$ 0.006} & \textbf{0.379 $\pm$ 0.028} & 1.530 $\pm$ 0.068 & 1.615 $\pm$ 0.053 & \textbf{0.270 $\pm$ 0.023} & $5\times 10^7$ \\
MD-Diff & \textbf{1.308 $\pm$ 0.072} & 0.113 $\pm$ 0.001 & 0.579 $\pm$ 0.004 & 3.627 $\pm$ 0.023 & 3.704 $\pm$ 0.026 & 0.310 $\pm$ 0.001 & $5\times 10^7$\\
MD-NF & 13.533 $\pm$ 0.024 & 0.138 $\pm$ 0.003 & 0.586 $\pm$ 0.003 & \textbf{0.551 $\pm$ 0.062} & \textbf{1.198 $\pm$ 0.069} & 0.403 $\pm$ 0.045 &   $5\times 10^7$  \\
TA-BG & 14.993 $\pm$ 0.002 & 0.219 $\pm$ 0.013 & 0.685 $\pm$ 0.034 & 83.457 $\pm$ 0.070 & 86.176 $\pm$ 0.104 & 0.979 $\pm$ 0.012 & $5\times 10^7$\\

FKC & 14.392 $\pm$ 0.909 & 0.217 $\pm$ 0.000 & 0.649 $\pm$ 0.001 & 11.281 $\pm$ 0.025 & 11.466 $\pm$ 0.027 & 2.120 $\pm$ 0.024 & $5\times 10^7$ \\

Score Scaling & 4.588 $\pm$ 0.467 & 0.183 $\pm$ 0.002 & 0.608 $\pm$ 0.008 & 10.282 $\pm$ 0.020 & 10.460 $\pm$ 0.019 & 0.550 $\pm$ 0.036 & $5\times 10^7$ \\
\bottomrule
\end{tabular}
}
\label{tab:aldp_tab}
\vspace{-10pt}
\end{table}

\begin{table}[t!]
\vspace{-5pt}
\caption{\small Performance of methods for the AL3 sampling task. The starting temperature is $T_L = 1200$ K, annealed to target $T_S = 300$K. Metrics are calculated over 10k samples and standard deviations over 3 seeds.}
\resizebox{\textwidth}{!}{%
\begin{tabular}{lrrrrrrr}
\toprule
  & Rama-KL & Tica-$\gW_1$ $\downarrow$ & Tica-$\gW_2$ $\downarrow$ & Energy-$\gW_1$ $\downarrow$ & Energy-$\gW_2$ $\downarrow$  & $\mathbb{T}$-$\mathcal{W}_2$ & \#Energy Evals \\
\midrule
\nameshort & \textbf{1.209 $\pm$ 0.144} & 0.272 $\pm$ 0.017 & 0.952 $\pm$ 0.055 & \textbf{2.567 $\pm$ 0.108} & \textbf{2.592 $\pm$ 0.107} & 0.521 $\pm$ 0.006 & $8 \times 10^7$\\
\nameshort (w/o relaxation) & 8.535 $\pm$ 0.254 & 0.405 $\pm$ 0.014 & 0.999 $\pm$ 0.043 & 86.270 $\pm$ 0.294 & 87.695 $\pm$ 0.294 & 0.651 $\pm$ 0.013 & $5 \times 10^7$ \\

MD-Diff & 9.662 $\pm$ 0.085& \textbf{0.059 $\pm$ 0.006} & \textbf{0.426 $\pm$ 0.010}  & 7.416 $\pm$ 0.130 & 7.599 $\pm$ 0.137 & 0.424 $\pm$ 0.011 & $8\times 10^7$\\

TA-BG  & 2.078 $\pm$ 2.088 & 0.082 $\pm$ 0.001 & 0.454 $\pm$ 0.001 & 4.782 $\pm$ 0.076 & 4.863 $\pm$ 0.082 & \textbf{0.347 $\pm$ 0.014} & $8\times 10^7$\\

TA-BG (w/o relaxation) & 14.988 $\pm$ 0.009 & 0.321 $\pm$ 0.001 & 0.648 $\pm$ 0.000 & 173.042 $\pm$ 0.717 & 178.558 $\pm$ 0.732 & 1.310 $\pm$ 0.004 & $5\times 10^7$\\

\bottomrule
\end{tabular}
}
\label{tab:al3_results}
\vspace{-10pt}
\end{table}

\begin{figure}[t!]
  \centering
  \includegraphics[width=\linewidth]{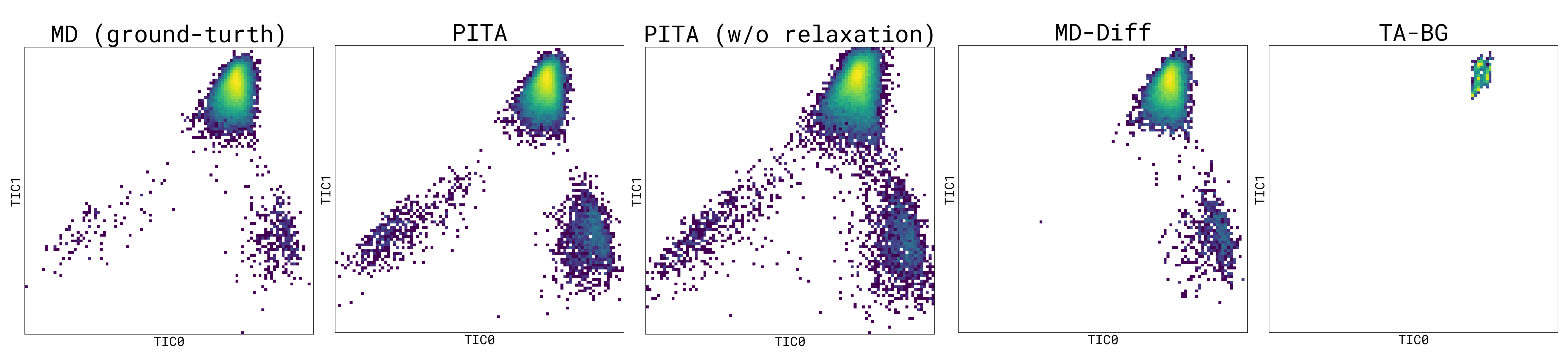}
  \caption{TICA plots for Alanine Tripeptide (AL3) at 300K obtained from different methods using 30k samples.}
  \label{fig:tica_al3}
\end{figure}

\paragraph{Alanine Tripeptide} We further evaluate the performance of \nameshort on a larger molecular system, Alanine Tripeptide (AL3). We employ a temperature annealing schedule with intermediate steps at ${1200\ \mathrm{K},\ 755.95\ \mathrm{K},\ 555.52\ \mathrm{K},\ 408.24\ \mathrm{K},\ 300\ \mathrm{K}}$. As shown in \cref{fig:tica_al3}, \nameshort again successfully recovers the essential dynamical modes of the system, indicating its capability of generating samples that align with the dominant kinetic features of the underlying dynamics. In practice, we also observe that performing a short additional MD refinement at the target temperature (300 K) after generation further improves the physical plausibility and smoothness of the trajectories. This leads to more accurate estimates of the free energy landscape, for both \nameshort and TA-. In \cref{tab:al3_results}, we provide quantitative analysis of the performance of the models. Notably, despite resulting in a better mode coverage, \nameshort performs worse than baselines according to Tica$-\gW_1$ and Tica$-\gW_2$, which suggest that it does not fully recover the correct relative weights of the modes.

%% file: text/conclusion.tex
\section{Conclusion}
\label{sec:conclusion}

\looseness=-1
In this paper, we propose \nameshort a new framework to train diffusion-based samplers by introducing two mechanisms of interpolating a target Boltzmann density by changing the temperature and defining a diffusion noising process. We demonstrated that \nameshort allows the progressive training of a sequence of diffusion models that go from high temperature, where ground truth data is simple to collect, to the lower temperature target temperature. Using \nameshort we demonstrated equilibrium sampling of $N$-body particle systems, and, for the first time, equilibrium sampling of alanine dipeptide and tripeptide in Cartesian coordinates. Importantly, we demonstrate \nameshort requires drastically fewer energy evaluations than existing diffusion samplers. 

We believe that \nameshort represents a step forward in the scalability of diffusion-based samplers and opens up ripe avenues for future work including improved training strategies and regimes for energy-based models that are in service of the \nameshort framework. Natural directions for future work include automatically determining the optimal temperature jump when instantiating our Feynman-Kac PDE to generate asymptotically unbiased samples at lower temperatures or transferable sampling~\citep{klein2024transferableboltzmanngenerators,tan2025amortizedsamplingtransferablenormalizing}.

\xhdr{Limitations} To obtain a consistent estimator or an importance sampling scheme, one has to define a density model of the generated samples. For this, \nameshort relies on training an additional energy-based model, which is a notoriously challenging task. Furthermore, simultaneous training and inference of both the score model and the energy-based model introduces additional computational and memory requirements.

%% file: text/acknowledgements.tex
\section*{Acknowledgments}

This project was partially sponsored by Google through the Google \& Mila projects program. The authors acknowledge funding from UNIQUE, CIFAR, NSERC, Intel, and Samsung. The research was
enabled in part by computational resources provided by the Digital Research Alliance of Canada (\url{https://alliancecan.ca}), Mila (\url{https://mila.quebec}), and NVIDIA. KN was supported by IVADO and Institut Courtois. AJB is partially supported by an NSERC Post-doc fellowship. This research is partially supported by the EPSRC Turing AI World-Leading Research Fellowship No. EP/X040062/1 and EPSRC AI Hub No. EP/Y028872/1.

%% file: text/appendix.tex
\section{Proofs}
\label{app:proofs}

\subsection{Inference-Time Annealing}
\label{app:annealing}
\begin{mdframed}[style=mybox]
\annealing*
\end{mdframed}
\begin{proof}
For the Energy-Based Model $U_t(x;\eta)$, we denote the corresponding Boltzmann density as
\begin{align}
    q_t(x) \propto \exp(-\gamma U_t(x;\eta))\,,
\end{align}
where $\gamma$ is the target inverse temperature. Taking the time-derivative of $q_t(x)$ we get the following equation
\begin{align}
    \deriv{q_t(x)}{t} =~& q_t(x)\left(-\gamma\deriv{U_{t}(x;\eta)}{t} + \mean_{q_t(x)}\gamma\deriv{U_{t}(x;\eta)}{t}\right)\,,
\end{align}
which can be simulated by reweighting the samples from $q_{t=0}(x) \approx \Normal(0,1)$ according to the following weights
\begin{align}
    w(x_1) = \frac{\exp(-\gamma\int_0^1 dt\; \partial U_{t}(x_t;\eta)/\partial t)}{\mean_{q_0(x_0)}\exp(-\gamma\int_0^1 dt\; \partial U_{t}(x_t;\eta)/\partial t)}\,.
\end{align}
Although this reasoning is theoretically justified, in practice, the variance of this importance-weighted estimate (or resampled distribution) is prohibitively large. That is why one has to introduce additional terms that move samples around tracing the original diffusion process. Namely, we consider the following PDE
\begin{align}
    \deriv{q_t(x)}{t} =~& \pm\inner{\nabla}{q_{t}(x)(-a_tx+\frac{\zeta_t^2}{2}s_{t}(x))} + q_t(x)\left(-\gamma\deriv{U_{t}(x;\eta)}{t} + \mean_{q_t(x)}\gamma\deriv{U_{t}(x;\eta)}{t}\right) \nonumber\\
    =~& -\inner{\nabla}{q_{t}(x)(-a_tx+\frac{\zeta_t^2}{2}s_t(x))} + q_{t}(x)\left(g_t(x) - \mean_{q_{t}(x)}g_t(x)\right)\,,\\
    g_t(x) =~& \gamma\inner{\nabla U_{t}(x;\eta)}{a_tx} - \gamma\frac{\zeta_t^2}{2}\inner{\nabla U_{t}(x;\eta)}{s_{t}(x)} - \inner{\nabla}{a_tx} + \\
    ~&+ \frac{\zeta_t^2}{2}\inner{\nabla}{s_{t}(x)} - \gamma\deriv{U_{t}(x;\eta)}{t}\,,
\end{align}
where the term $\inner{\nabla}{a_tx}$ does not depend on $x$ and cancels out when in the reweighting term. Furthermore, we can introduce the noise term by adding and subtracting the score $\xi_t(\zeta_t^2/2)\nabla\log q_t(x)$, i.e.
\begin{align}
    \deriv{q_t(x)}{t} =~& -\inner{\nabla}{q_{t}(x)(-a_tx+\frac{\zeta_t^2}{2}s_{t}(x) - \gamma\xi_t\frac{\zeta_t^2}{2}\nabla U_{t}(x;\eta) )} + \xi_t\frac{\zeta_t^2}{2}\Delta q_t(x) + \\
    ~&+ q_{t}(x)\left(g_t(x) - \mean_{q_{t}(x)}g_t(x)\right)\,,\\
    \begin{split}
    g_t(x) =~& -\gamma\inner{\nabla U_{t}(x;\eta)}{-a_tx + \frac{\zeta_t^2}{2}s_{t}(x)} + \frac{\zeta_t^2}{2}\inner{\nabla}{s_{t}(x)} - \gamma\deriv{U_{t}(x;\eta)}{t}\,,
    \label{appeq:reweight_func_annealing}
    \end{split}
\end{align}
which can be simulated as
\begin{align}
    dx_t =~& \left(-a_tx_t+\frac{\zeta_t^2}{2}\left(s_{t}(x_t) - \gamma\xi_t\nabla U_{t}(x_t;\eta)\right)\right)dt + \zeta_t\sqrt{\xi_t} dW_t\,,\\
    d\log w_t =~&  \left[-\gamma\inner{\nabla U_{t}(x_t;\eta)}{-a_tx_t + \frac{\zeta_t^2}{2}s_{t}(x_t)} + \frac{g_t^2}{2}\inner{\nabla}{s_{t}(x_t)} - \gamma\deriv{U_{t}(x_t;\eta)}{t}\right]dt\,.
\end{align}
\end{proof}

\begin{mdframed}[style=mybox]
\annealingvar*
\end{mdframed}
\begin{proof}
    Indeed, for $\gamma= 1$, \cref{appeq:reweight_func_annealing} becomes
    \begin{align}
        g_t(x) =~& -\inner{\nabla U_{t}(x;\eta)}{-a_tx + \frac{\zeta_t^2}{2}s_{t}(x)} + \frac{\zeta_t^2}{2}\inner{\nabla}{s_{t}(x)} - \deriv{U_{t}(x;\eta)}{t}\\
        =~& \inner{\nabla U_{t}(x;\eta)}{a_tx - \frac{\zeta_t^2}{2}s_{t}(x)} - \inner{\nabla}{a_tx_t - \frac{\zeta_t^2}{2}s_{t}(x)} + da_t - \deriv{U_{t}(x;\eta)}{t}\,,\nonumber
    \end{align}
    where $d$ is the dimensionality of the state-space. For $s_{t}(x) = -\nabla U_{t}(x;\eta) = \nabla\log p_{t}(x)$, we have
    \begin{align}
        g_t(x) =~& -\frac{1}{p_{t}(x)}\inner{\nabla}{p_{t}(x)\left(a_t x_t - \frac{\zeta_t^2}{2}\nabla\log p_{t}(x)\right)} +\deriv{\log p_{t}(x)}{t} + \deriv{\log Z_{t}}{t} + da_t\\
        =~& \frac{1}{p_{t}(x)} \underbrace{\left[-\inner{\nabla}{p_{t}(x)\left(a_t x- \frac{\zeta_t^2}{2}\nabla\log p_{t}(x)\right)} +\deriv{p_{t}(x)}{t}\right]}_{ = 0 \;\text{ due to \cref{eq:ref_FP}}} + \deriv{\log Z_{t}}{t} + da_t\,,
    \end{align}
    where $Z_{t} = \int dx\; p_{t}(x)$. The term in the brackets equals zero due to \cref{eq:ref_FP} since the ground true marginals $p_{t}(x)$ are defined as the marginals of the diffusion process. Hence, we have
    \begin{align}
        g_t(x) =~& \deriv{\log Z_{t}}{t} + da_t = \text{ constant of } x\,,
    \end{align}
    which becomes zero after the normalization $g_t(x) - \mean_{q_{t}(x)}g_t(x)$, which concludes the proof.
\end{proof}

\subsection{Inference-Time Geometric Averaging}
\label{app:averaging}

For the diffusion process with marginals $p_t(x)$ and the target distribution $p_{t=1}(x) \propto \Normal(x\cond 0,\one)^{(1-\beta_i)}\pi(x)^{\beta_i}$, we assume having the energy model $U_t(x;\eta)$ and the score model $s_t(x;\theta)$. Then, for the following density
\begin{align}
    q_t(x) \propto \exp\left(-\frac{\beta_{i+1}}{\beta_i}U_t(x;\eta) -\frac{\beta_{i+1}-\beta_i}{\beta_i}\log\Normal(x\cond 0,(\alpha_{1-t}^2 + \sigma_{1-t}^2)\one)\right)\,,
\end{align}
we have $q_{t=1}(x) \approx \Normal(x\cond 0,\one)^{(1-\beta_{i+1})}\pi(x)^{\beta_{i+1}}$. To sample from this density, we derive another SDE that performs inference-time geometric averaging. Analogously to \cref{prop:annealing}, for $\gamma = 1$ and perfectly trained models, the weights become constant, and this SDE yields the reverse-time diffusion SDE.

\begin{mdframed}[style=mybox]
\begin{restatable}{proposition}{averaging}
\label{prop:averaging}
{\textup{[Inference-time Geometric Averaging]}}
    For the geometric averaging of the energy-based model $q_t(x) \propto \exp\left((1-\gamma)(-U_{t,\beta}(x,\eta)) + \gamma \log\Normal(x\cond 0,\sigma_t^2)\right)$, the weighted samples from $q_t(x)$ can be collected by running the following SDE
    \begin{align}
        dx_t =~& -a_t x_t+(1-\gamma)\frac{\zeta_t^2}{2}(s_{t}(x_t)-\xi_t\nabla U_t(x_t;\eta)) - \gamma \frac{1}{\sigma_t^2}\left(1+\xi_t\frac{\zeta_t^2}{2}\right)x_t + \zeta_t\sqrt{\xi_t}dW_t\,,\nonumber\\
        d\log w_t =~& \inner{-(1-\gamma)\nabla U_t(x_t;\eta)-\gamma \frac{1}{\sigma_t^2}x_t}{-a_t x_t+(1-\gamma)\frac{\zeta_t^2}{2}s_{t}(x_t) - \gamma \frac{1}{\sigma_t^2}x_t} + \\
        ~&+(1-\gamma)\frac{\zeta_t^2}{2}\inner{\nabla}{s_{t}(x_t)} + (1-\gamma)\deriv{U_{t}(x_t;\eta)}{t} + \gamma \frac{1}{\sigma_t^3}\norm{x_t}^2\deriv{\sigma_t}{t}\,.
    \end{align}
    where $s_{t}(x)$ is any vector field. Finally, unweighted samples from $q_t(x)$ can be sampled using SNIS from \cref{eq:snis}.
\end{restatable}
\end{mdframed}
\begin{proof}
For the Energy-Based Model $U_t(x;\eta)$, we denote the corresponding geometric averaged density as
\begin{align}
    q_t(x) \propto \exp\left((1-\gamma)(-U_t(x;\eta)) + \gamma\log \Normal(x\cond 0,\sigma_{t}^2)\right)\,,
\end{align}
where $\gamma$ is the target inverse temperature. Taking the time-derivative of $q_t(x)$ we get the following equation
\begin{align}
    \deriv{q_t(x)}{t} =~& q_t(x)\left(g_t(x) - \mean_{q_t(x)}g_t(x)\right)\,,\\
    g_t(x) =~& (1-\gamma)\deriv{U_{t}(x;\eta)}{t} + \gamma \frac{1}{\sigma_t^3}\norm{x}^2\deriv{\sigma_t}{t}\,.
\end{align}
Assuming that the change of the density is close to the trained diffusion process, we introduce the drift-term corresponding to the score of the marginals
\begin{align}
    \deriv{q_t(x)}{t} =~& \underbrace{\pm\inner{\nabla}{q_{t}(x)(-a_t x+(1-\gamma)\frac{\zeta_t^2}{2}s_{t}(x) - \gamma \frac{1}{\sigma_t^2}x)}}_{\text{fictitious term}} +  q_t(x)\left(g_t(x) - \mean_{q_t(x)}g_t(x)\right)\,, \nonumber\\
    g_t(x) =~& (1-\gamma)\deriv{U_{t}(x;\eta)}{t} + \gamma \frac{1}{\sigma_t^3}\norm{x}^2\deriv{\sigma_t}{t}\,.
\end{align}
Moving the positive term to the weights and interpreting the negative term as the continuity equation, we get
\begin{align}
    \deriv{q_t(x)}{t} =~& -\inner{\nabla}{q_{t}(x)(-a_t x+(1-\gamma)\frac{\zeta_t^2}{2}s_{t}(x) - \gamma \frac{1}{\sigma_t^2}x)} +  q_t(x)\left(g_t(x) - \mean_{q_t(x)}g_t(x)\right)\,, \nonumber\\
    g_t(x) =~& \inner{-(1-\gamma)\nabla U_t(x;\eta)-\gamma \frac{1}{\sigma_t^2}x}{-a_t x+(1-\gamma)\frac{\zeta_t^2}{2}s_{t}(x) - \gamma \frac{1}{\sigma_t^2}x} + \\
    ~&+(1-\gamma)\frac{\zeta_t^2}{2}\inner{\nabla}{s_{t}(x)} + (1-\gamma)\deriv{U_{t}(x;\eta)}{t} + \gamma \frac{1}{\sigma_t^3}\norm{x}^2\deriv{\sigma_t}{t}\,.
\end{align}
Finally, we introduce the noise term by adding the drift 
\begin{align}
    \xi_t\frac{\zeta_t^2}{2}\nabla \log q_t(x) = \xi_t\frac{\zeta_t^2}{2}\left(-(1-\gamma)\nabla U_t(x;\eta)-\gamma\frac{1}{\sigma_t^2} x\right)\,.
\end{align}
Thus, we get
\begin{align}
    \deriv{q_t(x)}{t} =~& -\inner{\nabla}{q_{t}(x)\left(-a_t x+(1-\gamma)\frac{\zeta_t^2}{2}(s_{t}(x)-\xi_t\nabla U_t(x;\eta)) - \gamma \frac{1}{\sigma_t^2}\left(1+\xi_t\frac{\zeta_t^2}{2}\right)x\right)} + \nonumber\\
    ~& + \xi_t\frac{\zeta_t^2}{2}\Delta q_t(x) +  q_t(x)\left(g_t(x) - \mean_{q_t(x)}g_t(x)\right)\,,\\
    g_t(x) =~& \inner{-(1-\gamma)\nabla U_t(x;\eta)-\gamma \frac{1}{\sigma_t^2}x}{-a_t x+(1-\gamma)\frac{\zeta_t^2}{2}s_{t}(x) - \gamma \frac{1}{\sigma_t^2}x} + \\
    ~&+(1-\gamma)\frac{\zeta_t^2}{2}\inner{\nabla}{s_{t}(x)} + (1-\gamma)\deriv{U_{t}(x;\eta)}{t} + \gamma \frac{1}{\sigma_t^3}\norm{x}^2\deriv{\sigma_t}{t}\,.
\end{align}
The corresponding SDE is
\begin{align}
    dx_t =~& -a_t x_t+(1-\gamma)\frac{\zeta_t^2}{2}(s_{t}(x_t)-\xi_t\nabla U_t(x_t;\eta)) - \gamma \frac{1}{\sigma_t^2}\left(1+\xi_t\frac{\zeta_t^2}{2}\right)x_t + \zeta_t\sqrt{\xi_t}dW_t\,,\nonumber\\
    d\log w_t =~& \inner{-(1-\gamma)\nabla U_t(x_t;\eta)-\gamma \frac{1}{\sigma_t^2}x_t}{-a_t x_t+(1-\gamma)\frac{\zeta_t^2}{2}s_{t}(x_t) - \gamma \frac{1}{\sigma_t^2}x_t} + \\
    ~&+(1-\gamma)\frac{\zeta_t^2}{2}\inner{\nabla}{s_{t}(x_t)} + (1-\gamma)\deriv{U_{t}(x_t;\eta)}{t} + \gamma \frac{1}{\sigma_t^3}\norm{x_t}^2\deriv{\sigma_t}{t}\,.
\end{align}

\end{proof}

\section{Bridging the Gap at the End-Point}
\label{app:rendezvous}

Integrating the dynamics from \cref{prop:annealing,prop:averaging} we generate a set of weighted samples $\{(x^k_{t=1},w^k_{t=1})\}_{k=1}^K$ that converge to the samples from $q_{t=1}(x)$ when $K \to \infty$. In \cref{sec:inference} we assume that this density is defined as the Boltzmann distribution of the corresponding energy model, i.e. $q_{t=1}(x) \propto \exp(-\beta_{i+1}/\beta_i\cdot U_{t=1}(x;\eta))$, which approximates $\pi^{\beta_{i+1}}$, but does not necessarily match it exactly. Here we describe two possible ways to bridge the gap between the density model and the target density.

The first way to sample from $\pi(x)^{\beta_{i+1}}$ is via Self-Normalized Importance Sampling (SNIS). The integrated weights $w^k_{t=1} = e^{\int_0^1 dt\; g_t(x_t)}$ represent the density ratio between the distribution of the samples $x^k_{t=1}$ and the density of the integrated PDE (see discussion in \cref{sec:back_feynmankac}). Correspondingly, to sample from $\pi(x)^{\beta_{i+1}}$, we have to take into account the density ratio $\pi(x)^{\beta_{i+1}}/q_{t=1}(x)$ and obtain a new estimator, i.e.
\begin{align}
    \mean_{\pi(x)^{\beta_{i+1}}}\varphi(x) \propto~& \mean_{q_{1}(x)}\frac{\pi(x)^{\beta_{i+1}}}{q_1(x)}\varphi(x) \propto \mean \left[e^{\int_0^1 dt\; g_t(x_t)} \frac{\pi(x_1)^{\beta_{i+1}}}{q_1(x_1)}\varphi(x_1)\right]\,,\\
    \mean_{\pi(x)^{\beta_{i+1}}}\varphi(x) \approx~& \sum_{k=1}^K \frac{\tilde{w}_1^k}{\sum_{j=1}^n \tilde{w}_1^j}\varphi(x_1^k)\,,\; \tilde{w}_{1}^k \coloneqq e^{\int_0^1 dt\; g_t(x_t)}\pi(x^k_{1})^{\beta_{i+1}}/q_{1}(x^k_{1})\,,
\end{align}
where the new weights $\tilde{w}_1^k$ are obtained from the old ones $w_1^k$ by multiplication with the corresponding density ratio. Note, that the following empirical distribution approximates the target density $\pi(x)^{\beta_{i+1}}$
\begin{align}
    \tilde{\pi}(x)^{\beta_{i+1}} = \sum_{k=1}^K \frac{\tilde{w}_1^k}{\sum_{j=1}^n \tilde{w}_1^j} \delta(x-x_1^k)\,.
\end{align}

The alternative to importance sampling with the density model proposal is the gradual interpolation between the density model and the target during the integration. In particular, one can satisfy the boundary conditions by defining a smooth interpolant between the boundary densities $p_{t=0} = \Normal(0,I)$, $p_{t=1} = \pi^{\beta_{i+1}}$ and the annealed density model as follows
\begin{align}
    q_t(x) \propto \exp\bigg[~&-\gamma\left(1-\frac{t}{t_1}\right)_+^\kappa\log\Normal(0,1) - \left(1-\frac{1-t}{1-t_2}\right)_+^\kappa \beta_{i+1}\log \pi(x) -\\
    ~&- \left(1-\left(1-\frac{t}{t_1}\right)_+^\kappa - \left(1-\frac{1-t}{1-t_2}\right)_+^\kappa\right)\frac{\beta_{i+1}}{\beta_i} U_{t}(x;\eta)\bigg]\,,
\end{align}
where $(x)_+ = \max\{0,x\}$, $0 < t_1 < t_2 < 1$ are the hyperparameters that define switch times between models, and $\kappa$ is the smoothness parameter. Thus, we guarantee that $q_{t=1}(x) \propto \pi(x)^{\beta_{i+1}}$. However, in practice, we found that this interpolation technique results in a high variance of importance weights.

\section{Network Parameterization and Preconditioning}

\looseness=-1
We condition our score network $s$ and our energy network $U_t$ based on findings in EDM~\citep{karras2022elucidating}, use an energy parameterization based on \citet{neklyudov2023action} and \citet{thornton2025composition}, and include a new pre-conditioning on $\beta$. All of our networks are based on a backbone $F_\theta(x_t, t, \beta): (\R^d \times [0, \infty) \times [1, \infty)) \to \R^d$ is a flexible network architecture based on a diffusion transformer (DiT) backbone~\citep{peebles2023scalablediffusionmodelstransformers}. Specifically, we parameterize our denoiser network $D_\theta$ as:
\begin{equation}
    D_\theta(x_t, t, \beta) \coloneqq (1 +  \beta (c_{skip}(t) - 1) x_t + \beta c_{out}(t) F_\theta \left (c_{in}(t) x_t, c_{noise}(t) \right )
\end{equation}
which allows us to define our score network $s_\theta$ as
\begin{equation}
    s_\theta(x_t, t, \beta) := \frac{D_\theta(x_t, t, \beta) - x_t}{\sigma_t^2}
\end{equation}

We pre-condition the energy as

\begin{equation}
U_\eta(x_t, t, \beta) \coloneqq \beta \left ( \frac{1 - a_t c_{skip}(t)}{2 \sigma_t^2}  \norm{x_t}^2 - \frac{\xi_t c_{out}(t)}{c_{in}(t) \sigma_t^2} \left ( x_t \cdot F_\eta(c_{in}(t) x_t, c_{noise}(t))\right )\right )
\end{equation}

\section{Molecular Dynamics Analysis}
\label{app:md-mixing}
\subsection{MD mode mixing across temperatures}
\begin{figure}[h]
\vspace{-10pt}
  \centering
  \begin{subfigure}[t]{\linewidth}
    \centering
    \includegraphics[width=0.95\linewidth]{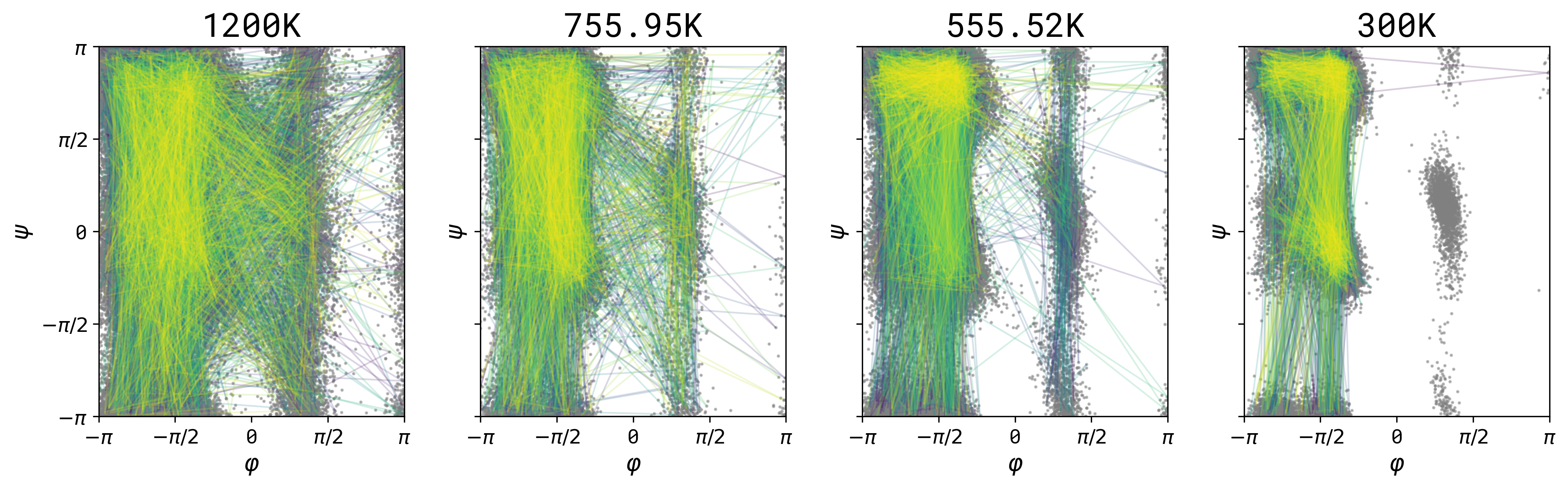}
    \caption{Visualization of the Ramachandran plots of the MD chain over time, the lines are colored from purple to yellow over 50 million MD steps. The gray plot shows the Ramachandran plot of a chain with 1 billion MD steps. } 
    \label{fig:}
  \end{subfigure}
  \vspace{1em} 
  \begin{subfigure}[t]{\linewidth}
    \centering
    \includegraphics[width=0.95\linewidth]{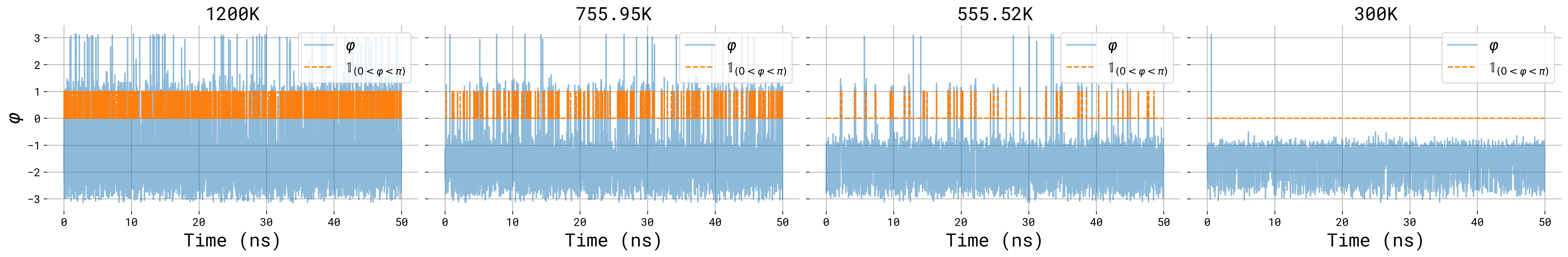}
    \caption{Trace plot for the internal angle $\phi$. The blue lines show the value of $\phi$ across the 50-million-step MD chain. The orange line indicates when the chain switches out of the main modes. }
    \label{fig:}
  \end{subfigure}
\vspace{-10pt}
  \caption{Analysis of the mixing of the MD chains for ALDP, for a 50 million-step MD simulation, across the annealing temperatures.}
  \label{fig:alp_md_analysis}
\end{figure}

\begin{figure}[h]
\vspace{-10pt}
  \centering
  \begin{subfigure}[t]{\linewidth}
    \centering
    \includegraphics[width=0.95\linewidth]{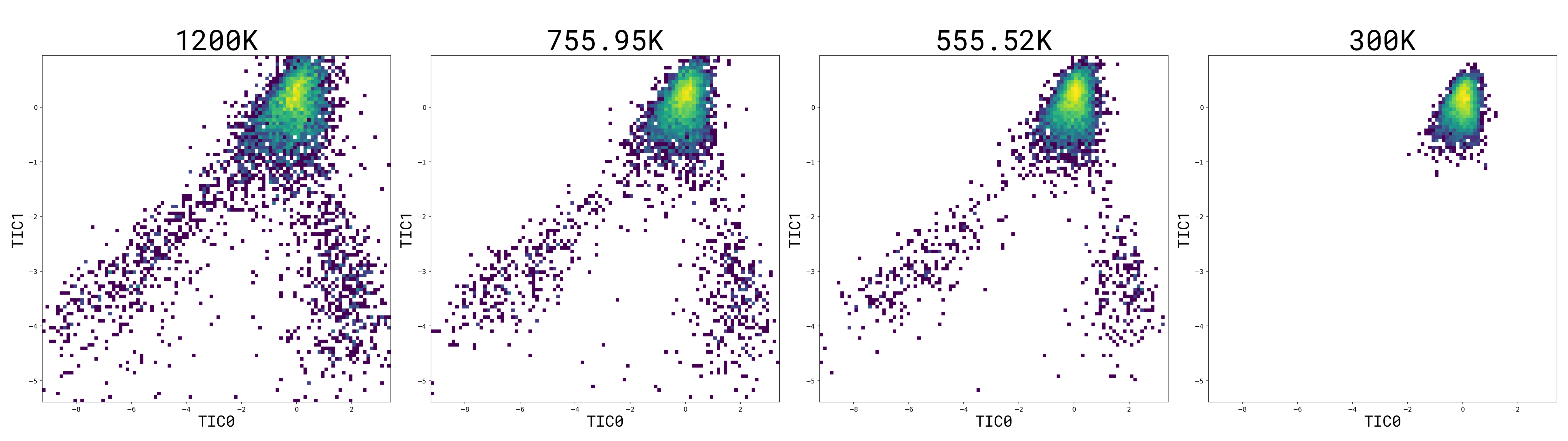}
    \caption{AL3 TICA plots of ground-truth MD samples across different temperatures, using MD chains of 50 million steps. For all temperatures, the TICA axes are matched to those of 300K.}
    \label{fig:}
  \end{subfigure}
  \vspace{1em} 
  \begin{subfigure}[t]{\linewidth}
    \centering
    \includegraphics[width=0.95\linewidth]{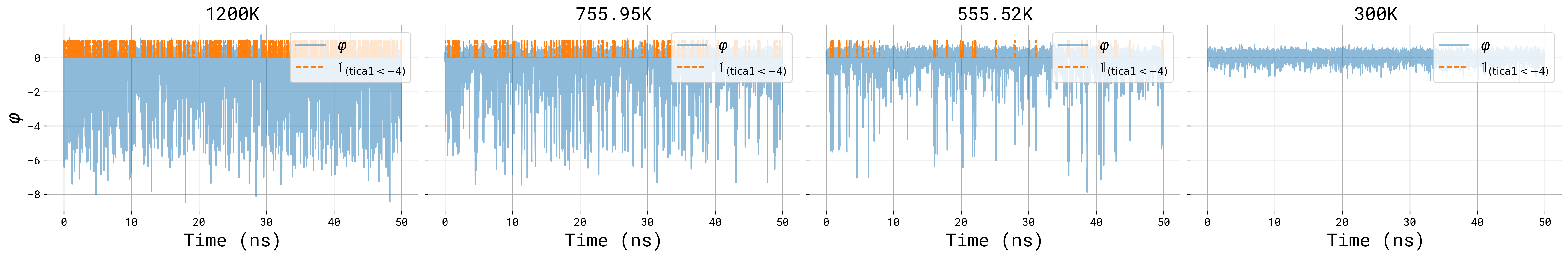}
    \caption{Trace plot for the second TICA axis of ground-truth MD samples. The blue lines show the value of the second TICA axis across the 50-million-step MD chain. The orange line indicates when the chain switches out of the main mode. }
    \label{fig:}
  \end{subfigure}
\vspace{-10pt}
  \caption{Analysis of the mixing of the MD chains for AL3, for a 50 million-step MD simulation, across the annealing temperatures. }
  \label{fig:al3_md_analysis}
\end{figure}

In ~\cref{fig:alp_md_analysis} and~\cref{fig:al3_md_analysis}, we analyze the mixing behaviour of MD simulations for ALDP and AL3 across various annealing temperatures. Specifically, we examine simulations consisting of 50 million steps—matching the quantity of MD data used for training \nameshort\ at 1200K. As the temperature decreases, the sampling quality deteriorates: the chains exhibit poorer mixing and fail to explore significant regions of the configuration space, missing major modes of the distribution. This is shown both in Ramachandran and TICA plots, as well as the trace plots of the internal angle $\phi$ and the second TICA axis. More specifically, for ALDP, we see that the chain switches out of the main mode $5.8\%$, $3.0\%$, $1.0\%$ and $0 \%$ of the time at temperatures 1200K, 755.95K, 555.52K and 300K, respectively. For AL3, this happens at a rate of $12.7\%$, $9.2\%$, $5.8\%$ and $0 \%$.

This motivates training at a higher temperature then annealing to a lower temperature as is done in \nameshort. As we are able to take advantage of relatively quick mode mixing at higher temperatures and the ability of inference time annealing to recover samples from a lower temperature. 

\vspace{-10pt}

\subsection{Temperature Annealing ESS} \label{app:temp_ess}
\begin{figure}
    \centering
    \includegraphics[width=0.5\linewidth]{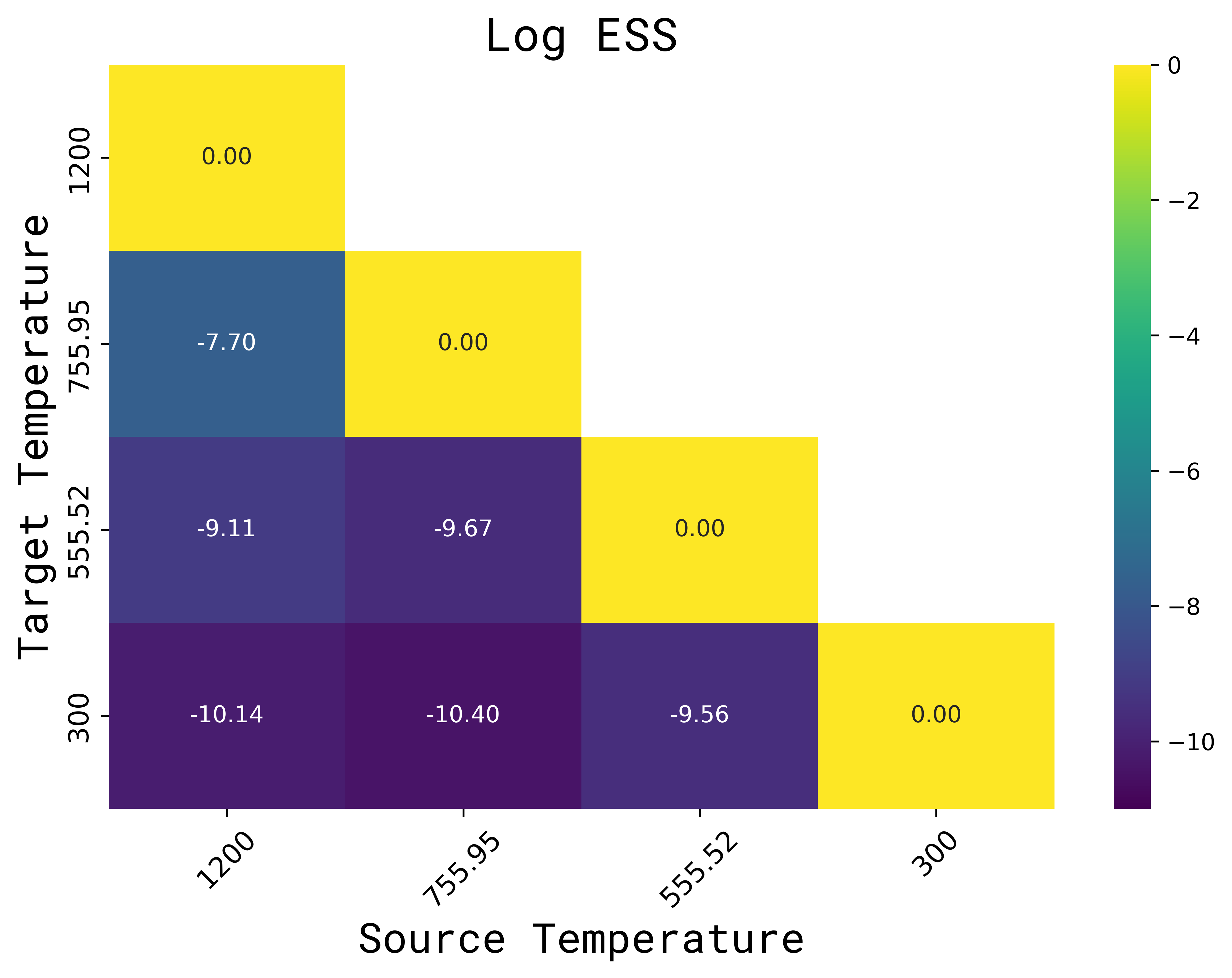}
    \caption{Log effective sample size (ESS) values for importance sampling from different source and target temperatures for ALDP. The ESS values are computed on 33000 samples from the MD chain. The low values indicate that purely relying on importance sampling is not sufficient to sample from lower temperature targets.}
    \label{fig:ess}
\end{figure}

\begin{figure}
    \centering
    \includegraphics[width=0.5\linewidth]{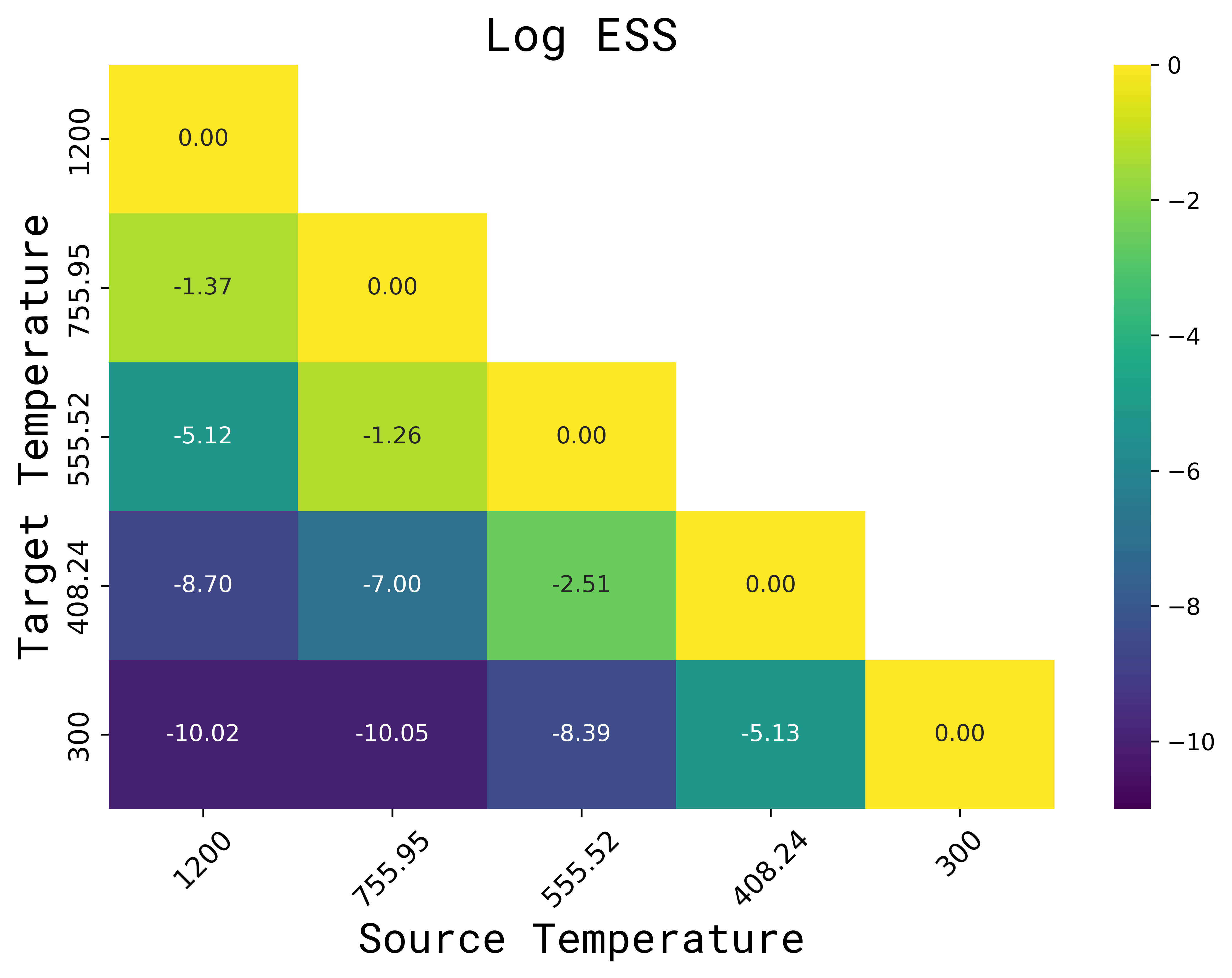}
    \caption{Log effective sample size (ESS) values for importance sampling from different source and target temperatures for AL3.  The ESS values are computed on 33000 samples from the MD chain. The low values indicate that purely relying on importance sampling is not sufficient to sample from lower temperature targets.}
    \label{fig:ess_al3}
\end{figure}

To show the difficulty of directly performing importance sampling directly across large temperature jumps in coordinate space using methods similar to \citet{schopmans2025temperatureannealedboltzmanngenerators,rissanen2025progressivetemperingsamplerdiffusion} we calculate the ESS assuming these models could learn the source temperature perfectly. To perform this experiment we take 33k samples from the test chain and calculate the (normalized) effective sample size of these points when importance sampled to a lower target energy. The results of this experiment are presented in \cref{fig:ess} for ALDP and \cref{fig:ess_al3} for AL3. Here we report the Log normalized effective sample size for each temperature jump considered in this paper. We compute the log normalized ESS using Kish's formula normalized by the number of samples as:
\begin{equation}
\text{log ESS}\left(\{w_i\}_{i=1}^N\right) = \log\frac{\left( \sum_{i=1}^{N} w_i \right)^2}{N\sum_{i=1}^{N} w_i^2}.
\end{equation}

Where $w_i$ is the importance weight of the $i$th sample. We note that a log ESS of $-5$ would mean that direct importance sampling would require $100\,000$ times as many points at the higher temperature as are needed at the lower temperature. This makes importance sampling computationally infeasible for these temperature jumps, even with perfectly learned models.

This motivates the guided approach taken in \nameshort, where we are able to guide the samples preferentially towards the target temperature, avoiding the problem of low ESS between these temperature jumps.

\section{Additional Baselines}
\label{app:additional_baselines}
For the ALDP experiment, we compare against two additional baselines of annealed MCMC with sequential Monte Carlo (SMC) and parallel tempering (PT) strategies. For SMC, we take 10 annealing temperatures between 1200\,K and 300\,K using a geometric schedule, taking 166 steps per temperature with 30k particles. For PT, we use the same 4 annealing temperatures as PITA, running 25 MD steps between particle exchanges for a total of 50k iterations. As it can be seen in \cref{tab:aldp_smc_replica}, PITA performs better overall in capturing the distribution given the same budget of energy evaluation. In \cref{tab:aldp_inference_time}, we also compare the wall-clock time for inference on 30K samples for a trained PITA model as well as classical baselines, SMC and PT. The training time for PITA on ALDP is 564.3 minutes. We note that PITA is roughly an order of magnitude faster than PT and two orders of magnitude faster than SMC after amortization. Therefore, the wall time breakeven point is around 120k samples vs. SMC and 300k samples vs PT for PITA.

\begin{table}[t]
\centering
\caption{\small Performance of the SMC and PT baselines on ALDP}
\label{tab:aldp_smc_replica}
\resizebox{\textwidth}{!}{%
\begin{tabular}{lccccccc}
\toprule
  & Rama-KL & Tica-$\gW_1$ $\downarrow$ & Tica-$\gW_2$ $\downarrow$ & Energy-$\gW_1$ $\downarrow$ & Energy-$\gW_2$ $\downarrow$  & $\mathbb{T}$-$\mathcal{W}_2$ & \#Energy Evals \\
\midrule
\nameshort &
\textbf{$4.773 \pm 0.460$ }&
\textbf{$0.112 \pm 0.006$ }&
\textbf{$0.379 \pm 0.028$ }&
\textbf{$1.530 \pm 0.068$ }&
\textbf{$1.615 \pm 0.053$ }&
\textbf{$0.270 \pm 0.023$ }&
\textbf{$5\times 10^{7}$ }\\
PT &
$7.306 \pm 1.077$ &
$0.625 \pm 0.010$ &
$0.895 \pm 0.016$ &
$4.652 \pm 0.015$ &
$4.689 \pm 0.014$ &
$0.911 \pm 0.004$ &
$5\times 10^{7}$ \\
SMC &
$5.935 \pm 0.228$ &
$0.372 \pm 0.006$ &
$0.425 \pm 0.003$ &
$0.969 \pm 0.078$ &
$1.002 \pm 0.072$ &
$0.874 \pm 0.016$ &
$5\times 10^{7}$ \\
\bottomrule
\end{tabular}
}
\end{table}

\begin{table}[h!]
\centering
\caption{\small Inference time for 30k samples on ALDP}
\label{tab:aldp_inference_time}
\begin{tabular}{l c}
\hline
 & Time (min) \\
\hline
PITA & 4.7 \\
SMC & 139.2 \\
PT & 59.5 \\
\hline
\end{tabular}
\end{table}

\section{Additional Results}
\label{app:additional_results}
\begin{figure}
    \centering
    \includegraphics[width=0.5\linewidth]{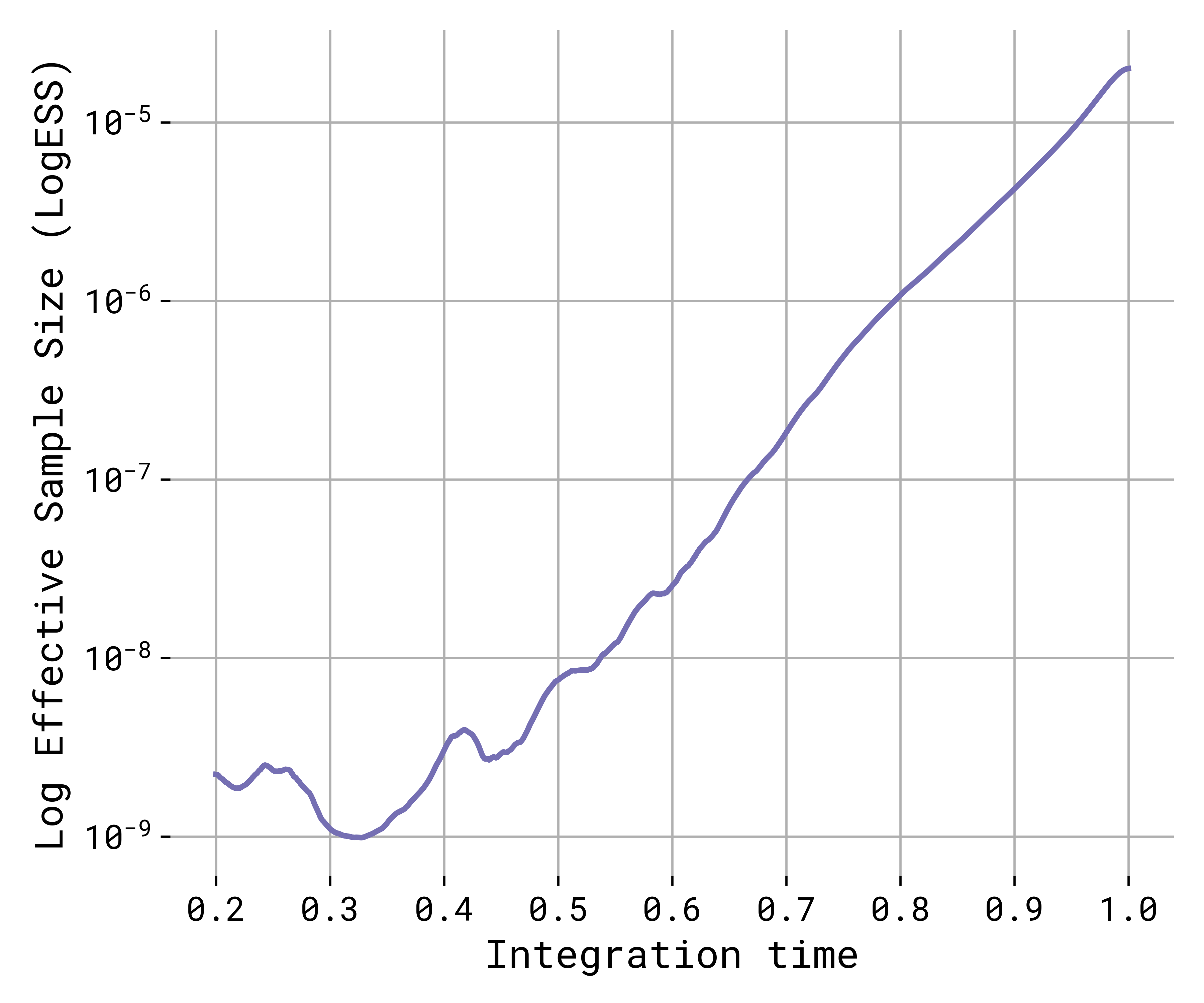}
    \caption{Log effective sample size (ESS) over integration time during inference.}
    \label{fig:ess_over_integration}
\end{figure}

In \cref{tab:aldp_ess_ais}, we report the Effective Sample Size (ESS) during \nameshort inference when importance weights are accumulated without SMC-based resampling on the ALDP dataset. The values are computed using 30000 samples for each of the temperature annealing steps. Note that these results correspond to a purely importance sampling-based setting (AIS) and thus differ from the full SMC formulation employed in PITA. The relatively low ESS values with AIS underscore the necessity of resampling for maintaining particle diversity and numerical stability. Additionally, we visualize the ESS as a function of the integration time during inference in \cref{fig:ess_over_integration}. The results indicate that most of the ESS variation occurs at the early stages $t<0.5$ (i.e., at higher noise levels), after which the ESS stabilizes. These findings suggest possible avenues for future work, such as exploring drift modifications aimed at reducing the variance of the importance weights.

\begin{table}[h!]
\centering
\caption{\small Log effective sample size (ESS) values without resampling for each temperature annealing step calculated across 30k samples on ALDP}
\label{tab:aldp_ess_ais}
\begin{tabular}{l c}
\hline
$T_L$ to $T_S$ & ESS ($\times 10^{-4}$) \\
\hline
1200K to 755.95K & 2.77 \\
755.95K to 555.52K & 8.66 \\
555.52K to 300K & 3.79 \\
\hline
\end{tabular}
\end{table}

\section{Ablation Studies}
\label{app:additional_ablations}
To evaluate the impact of our design choices, we perform a series of ablation studies examining: (1) the effect of annealing to 300K using different temperature jump sizes, (2) the choice of the $\gamma_t$ schedule, and (3) the role of resampling and different loss components, as well as comparing out method with a simple classifier-free guidance approach. The temperature jump size ablation is performed for both ALDP and AL3, while the remaining studies are conducted on ALDP.

\begin{figure}[t]
  \centering
  \begin{subfigure}[t]{\linewidth}
    \centering
    \includegraphics[width=0.9\linewidth]{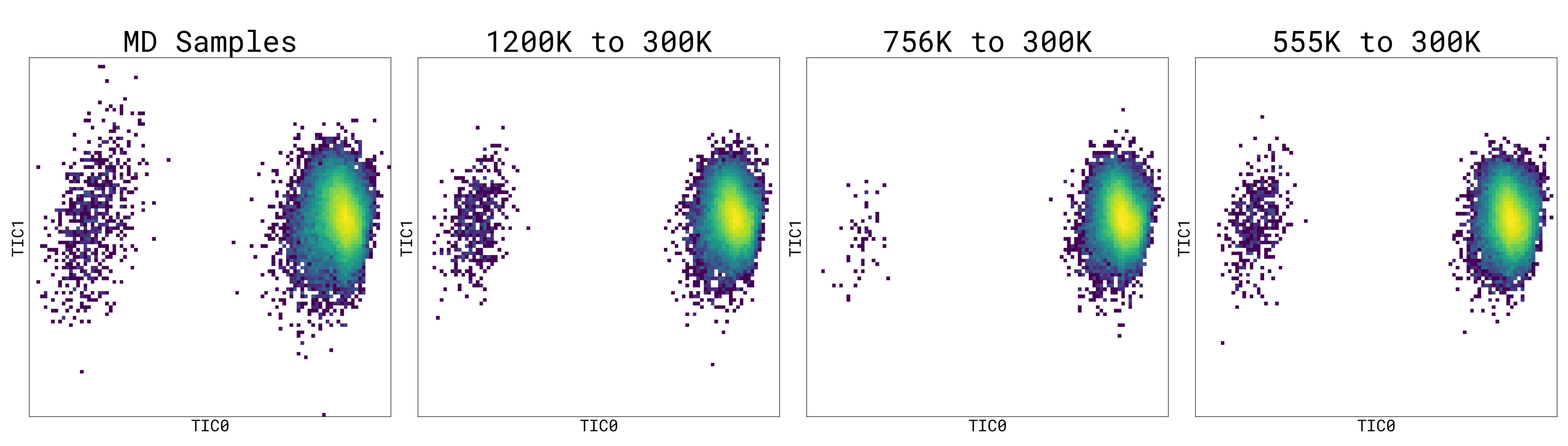}
    \caption{Alanine Dipeptide}
    \label{fig:aldp_jump_tica}
  \end{subfigure}

  \begin{subfigure}[t]{\linewidth}
    \centering
    \includegraphics[width=0.9\linewidth]{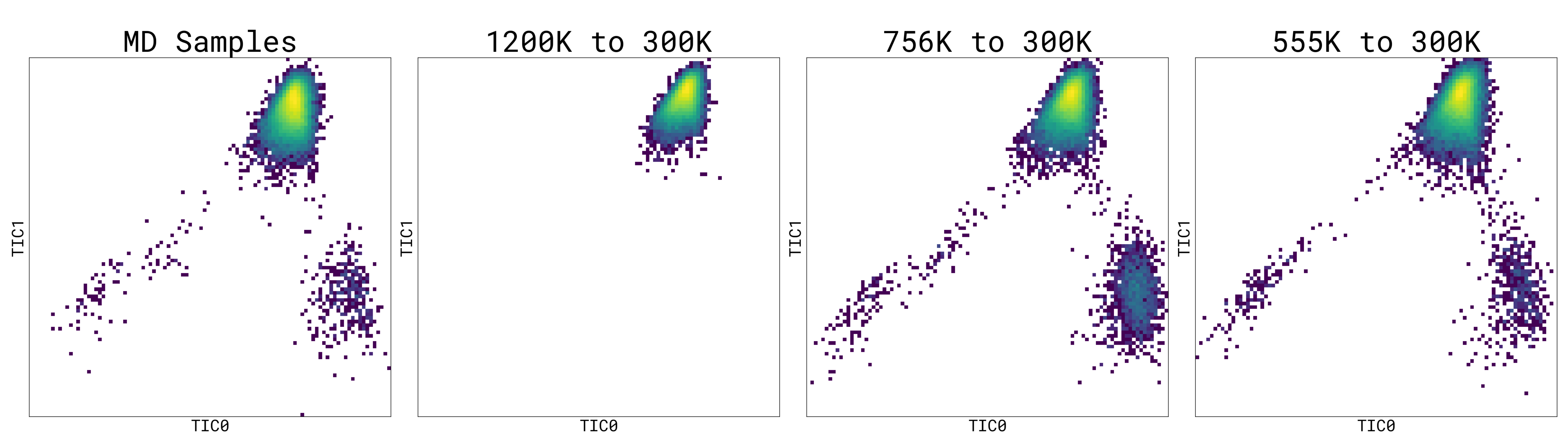}
    \caption{Alanine Tripeptide}
    \label{fig:al3_jump_tica}
  \end{subfigure}
  \caption{TICA plot of ALDP and AL3 samples obtained via annealing from various starting temperatures to $300$K.}
  \label{fig:alp_jump_tica_comparison}
\vspace{-10pt}
\end{figure}

\begin{figure}[t]
\vspace{-5pt}
  \centering
  \begin{subfigure}[t]{0.45\linewidth}
    \centering
    \includegraphics[width=\linewidth]{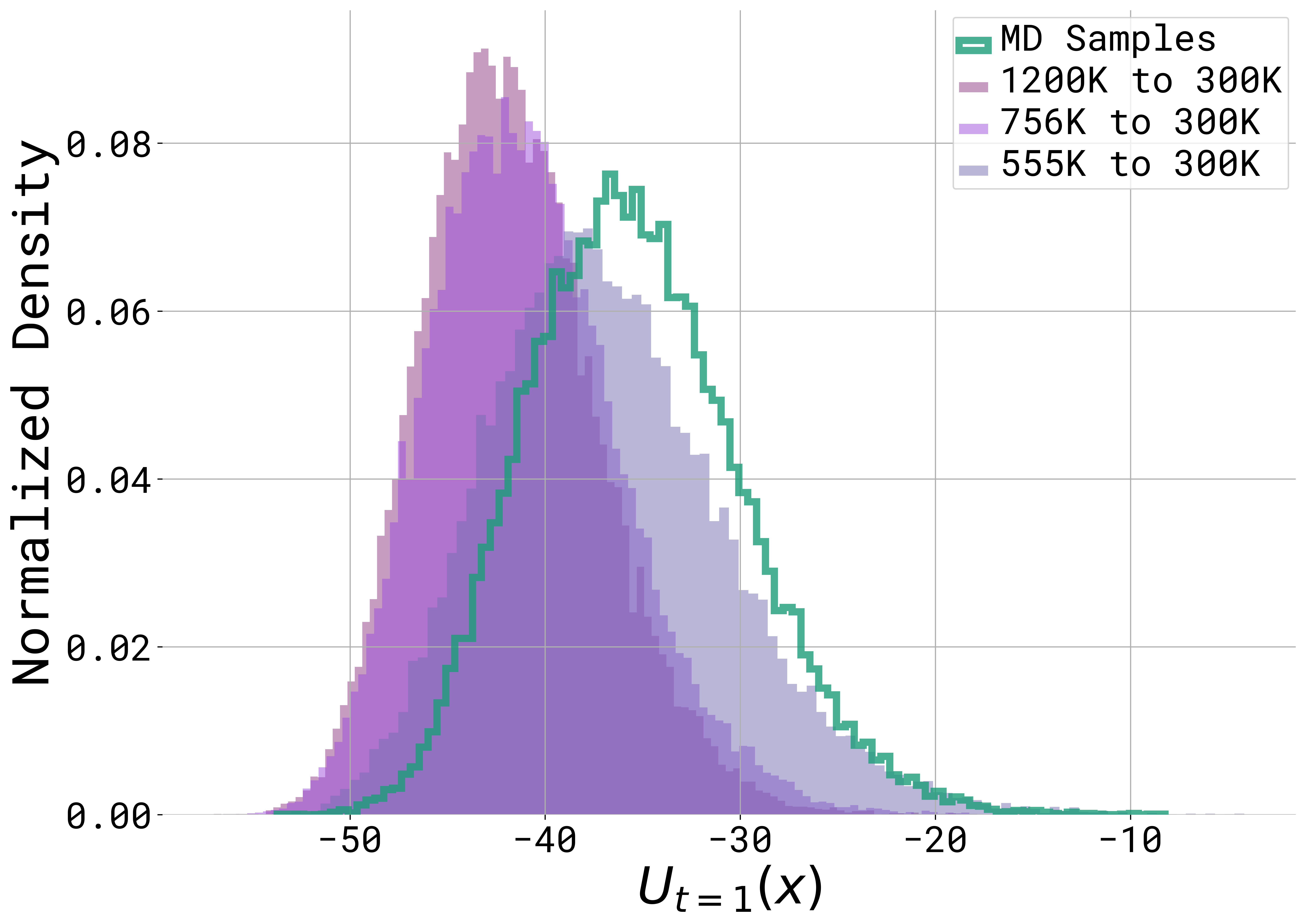}
    \caption{Alanine Dipeptide}
    \label{fig:aldp_jump_energy}
  \end{subfigure}
  \hfill
  \begin{subfigure}[t]{0.45\linewidth}
    \centering
    \includegraphics[width=\linewidth]{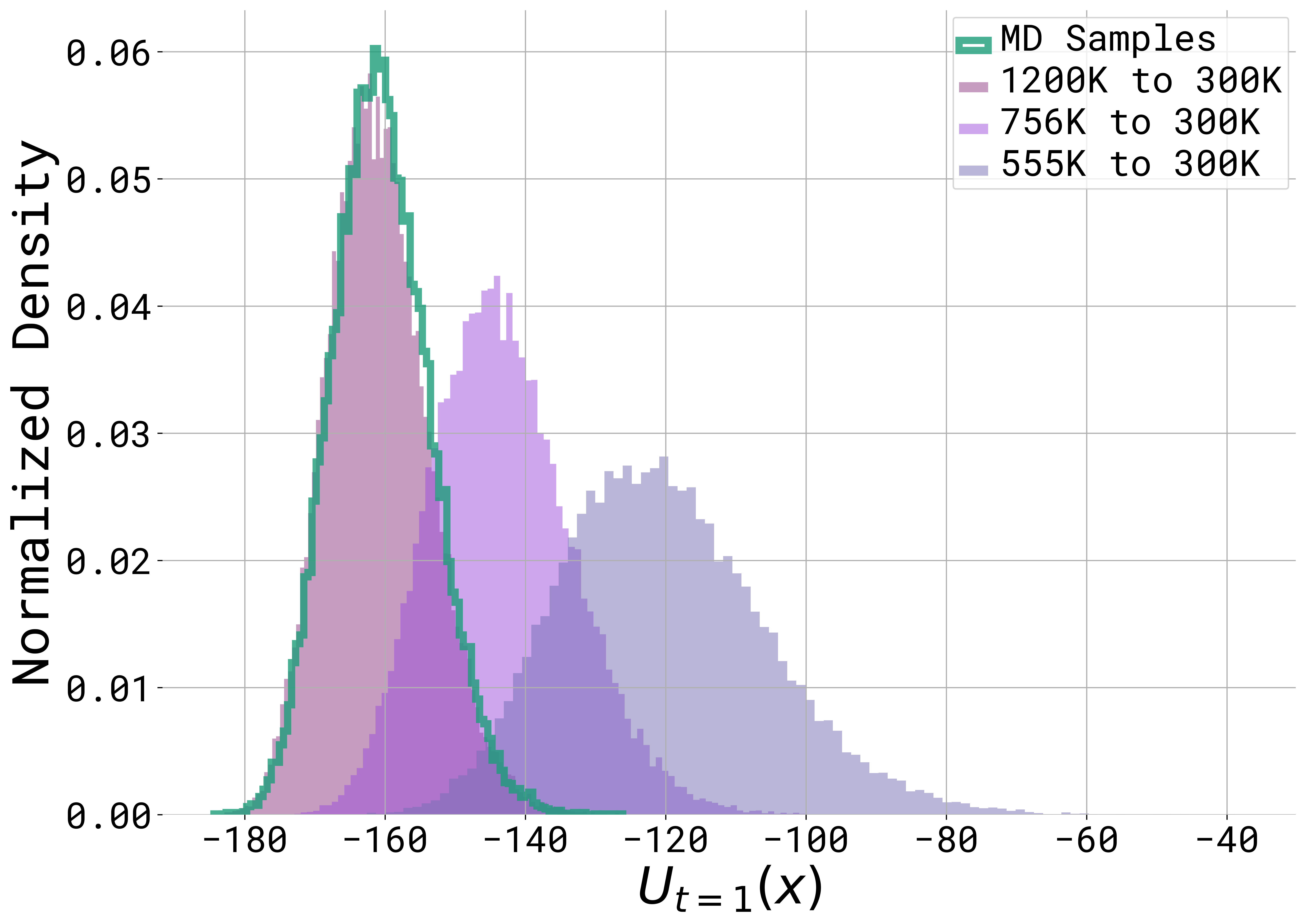}
    \caption{Alanine Tripeptide}
    \label{fig:al3_jump_energy}
  \end{subfigure}
  \caption{Energy distributions of ALDP and AL3 samples obtained via annealing from various starting temperatures to $300$K. The model is trained on ground-truth MD samples at $1200$K, and on annealed intermediate samples at $755.95$K and $555.52$K — the latter being our default training setting.}
  \label{fig:alp_jump_energy_comparison}
\end{figure}

\xhdr{Temperature Jump Sizes in Annealing} 
To evaluate the effectiveness of the progressive annealing schedule, we compare the performance of models where the system is annealed from different starting temperatures to $300$K. For ALDP, skipping intermediate temperatures has the most pronounced impact on energy distribution metrics, as shown in \cref{tab:aldp_jump} and \cref{fig:aldp_jump_energy}. In the case of AL3, we find that sequential training is essential for reliably capturing all modes at the lower temperature, as illustrated in \cref{fig:al3_jump_tica}. However, energy metrics degrade, likely due to small deviations in the sampled buffers at each annealing step, which accumulate over time. Accurately capturing the energy distribution in the sampled buffers at each intermediate temperature appears to be more difficult for AL3, which may contribute to the observed degradation. Nonetheless, capturing the correct modes remains a key priority, as modes lost during training are difficult to recover later. In contrast, mild deviations in the energy distribution can often be corrected through short MD relaxation steps, as demonstrated in \cref{sec:main_results}.

\begin{table}[t]
\vspace{-5pt}
\caption{Effect of different starting temperatures on annealing performance for ALDP and AL3, evaluated at the final temperature of $300$K.}
\centering
\begin{subtable}[t]{\textwidth}
\caption{ALDP.}
\resizebox{\textwidth}{!}{
\begin{tabular}{llllll}
\toprule
 & Tica-$\gW_1$ $\downarrow$ & Tica-$\gW_2$ $\downarrow$ & Energy-$\gW_1$ $\downarrow$ & Energy-$\gW_2$ $\downarrow$ & $\mathbb{T}$-$\mathcal{W}_2$ $\downarrow$ \\
$T_L$ to $T_S$ &  &  &  &  &  \\
\midrule
1200K to 300K & \textbf{0.100 $\pm$ 0.004} & \textbf{0.297 $\pm$ 0.019} & 6.438 $\pm$ 0.024 & 6.531 $\pm$ 0.021 & 0.301 $\pm$ 0.023 \\
755.95K to 300K & 0.180 $\pm$ 0.002 & 0.611 $\pm$ 0.003 & 5.639 $\pm$ 0.072 & 5.683 $\pm$ 0.070 & 0.358 $\pm$ 0.018 \\
555.52K to 300K & 0.121 $\pm$ 0.004 & 0.404 $\pm$ 0.019 & \textbf{1.541 $\pm$ 0.009} & \textbf{1.619 $\pm$ 0.010} & \textbf{0.270 $\pm$ 0.023} \\
\bottomrule
\end{tabular}
}
\label{tab:aldp_jump}
\end{subtable}

\vspace{0.1em} 

\begin{subtable}[t]{\textwidth}
\caption{AL3.}
\resizebox{\textwidth}{!}{
\begin{tabular}{llllll}
\toprule
 & Tica-$\gW_1$ $\downarrow$ & Tica-$\gW_2$ $\downarrow$ & Energy-$\gW_1$ $\downarrow$ & Energy-$\gW_2$ $\downarrow$ & $\mathbb{T}$-$\mathcal{W}_2$ $\downarrow$ \\
$T_L$ to $T_S$ &  &  &  &  &  \\
\midrule
1200K to 300K & 0.291 $\pm$ 0.005 & 0.558 $\pm$ 0.003 & \textbf{0.521 $\pm$ 0.122} & \textbf{0.597 $\pm$ 0.110} & 1.351 $\pm$ 0.014 \\
755.95K to 300K & 0.234 $\pm$ 0.009 & 0.663 $\pm$ 0.019 & 17.147 $\pm$ 0.105 & 17.429 $\pm$ 0.107 & 0.751 $\pm$ 0.006 \\
555.52K to 300K & \textbf{0.158 $\pm$ 0.004} & \textbf{0.329 $\pm$ 0.025} & 40.222 $\pm$ 0.198 & 40.978 $\pm$ 0.208 & \textbf{0.621 $\pm$ 0.038} \\
\bottomrule
\end{tabular}
}
\label{tab:al3_jump}
\end{subtable}
\label{tab:alp_jump}
\end{table}

\xhdr{$\gamma_t$ schedule}
We analyze the effects of using different schedules for time-dependent $\gamma_t$ during inference. In particular, we annneal from 555.52K to 300.0K using a constant schedule, a linear schedule which linearly increases from $\gamma=1$ to $\gamma=1.85$, and a sigmoid schedule again, increasing from $\gamma=1$ to $\gamma=1.85$. \cref{tab:aldp_gamma} shows that the linear schedule generally performs best across the different evaluation metrics. It achieves the lowest values on the TICA metrics, while showing comparable performance to the sigmoid schedule on the energy-based metrics. %

\xhdr{Resampling, Energy Pinning Loss, and Classifier Free Guidance}
To evaluate the impact of resampling on sample quality, we perform inference from $556$K to $300$K without applying resampling. Across all metrics, we observe that resampling consistently improves performance. To assess the roles of the energy pinning loss and classifier-free guidance, we retrain a model for the $556$K to $300$K transition with each component removed. Omitting the energy pinning loss results in a slight improvement in TICA metrics but leads to a noticeable decline in energy metrics, indicating that the loss plays an important role in maintaining accurate energy distributions. Finally, we train a diffusion model on the data generated from \nameshort at 555.52K, and anneal to 300K simply by scaling the score by $\gamma$ (similar to classifier-free guidance approaches). This approach shows mixed results, offering no consistent improvement over MD-Diff (which is directly trained on samples at 300K) and performing below the level of \nameshort in all metrics.

\begin{table}[H]
\vspace{-10pt}
\caption{$\gamma_t$ schedule ablation on ALDP.}
\resizebox{\textwidth}{!}{
\begin{tabular}{llllll}
\toprule
 & Tica-$\gW_1$ $\downarrow$ & Tica-$\gW_2$ $\downarrow$ &Energy-$\gW_1$ $\downarrow$ & Energy-$\gW_2$ $\downarrow$ & $\mathbb{T}$-$\mathcal{W}_2$ $\downarrow$ \\
$\gamma$ Schedule &  &  &  &  &  \\
\midrule
Constant & 0.115 $\pm$ 0.005 & 0.389 $\pm$ 0.014 & 1.485 $\pm$ 0.100 & 1.584 $\pm$ 0.095 & \textbf{0.258 $\pm$ 0.030} \\
Linear & \textbf{0.095 $\pm$ 0.009} & \textbf{0.243 $\pm$ 0.048} & 1.453 $\pm$ 0.099 & 1.555 $\pm$ 0.099 & 0.275 $\pm$ 0.058 \\
Sigmoid & 0.113 $\pm$ 0.009 & 0.339 $\pm$ 0.046 & \textbf{1.443 $\pm$ 0.087} & \textbf{1.550 $\pm$ 0.087} & 0.345 $\pm$ 0.027 \\
\bottomrule
\end{tabular}
}
\label{tab:aldp_gamma}
\end{table}

\begin{table}[H]
\vspace{-10pt}
\caption{Additional Ablation Results on ALDP.}
\resizebox{\textwidth}{!}{
\begin{tabular}{llllll}
\toprule
 & Tica-$\gW_1$ $\downarrow$ & Tica-$\gW_2$ $\downarrow$ &Energy-$\gW_1$ $\downarrow$ & Energy-$\gW_2$ $\downarrow$ & $\mathbb{T}$-$\mathcal{W}_2$ $\downarrow$ \\
\midrule
\nameshort & 0.121 $\pm$ 0.004 & 0.404 $\pm$ 0.019 & \textbf{1.541 $\pm$ 0.009} & \textbf{1.619 $\pm$ 0.010} & 0.270 $\pm$ 0.023 \\
w/o resampling & 0.140 $\pm$ 0.007 & 0.452 $\pm$ 0.027 & 1.606 $\pm$ 0.094 & 1.676 $\pm$ 0.075 & 0.363 $\pm$ 0.023 \\
w/o energy pinning loss & \textbf{0.098 $\pm$ 0.012} & \textbf{0.291 $\pm$ 0.065} & 4.709 $\pm$ 0.091 & 4.722 $\pm$ 0.090 & \textbf{0.219 $\pm$ 0.021} \\
MD-Diff + CFG & 0.137 $\pm$ 0.007 & 0.446 $\pm$ 0.029 & 8.106 $\pm$ 0.025 & 8.190 $\pm$ 0.026 & 0.383 $\pm$ 0.042 \\
\bottomrule
\end{tabular}
}
\end{table}
\label{tab:aldp_ablations_more}

\begin{table}[H]
\vspace{-10pt}
\caption{Metrics across temperatures}
\centering
\resizebox{\textwidth}{!}{
\begin{tabular}{llccccc}
\toprule
Temperature & Model & Tica-$\gW_1$ $\downarrow$ & Tica-$\gW_2$ $\downarrow$ & Energy-$\gW_1$ $\downarrow$ & Energy-$\gW_2$ $\downarrow$ & $\mathbb{T}$-$\mathcal{W}_2$ $\downarrow$ \\
\midrule
\multirow{2}{*}{755.95K} 
  & \nameshort & \textbf{0.024 $\pm$ 0.004} & \textbf{0.125 $\pm$ 0.021} & 2.855 $\pm$ 0.083 & 2.886 $\pm$ 0.079 & \textbf{0.134 $\pm$ 0.013} \\
  & TA-BG      & 0.040 $\pm$ 0.002 & 0.178 $\pm$ 0.008 & \textbf{2.065 $\pm$ 0.044} & \textbf{2.140 $\pm$ 0.042} & 0.355 $\pm$ 0.007 \\
\midrule
\multirow{2}{*}{555.52K} 
  & \nameshort & \textbf{0.141 $\pm$ 0.001} & \textbf{0.836 $\pm$ 0.004} & \textbf{1.420 $\pm$ 0.030} & \textbf{1.430 $\pm$ 0.027} & \textbf{0.219 $\pm$ 0.015} \\
  & TA-BG      & 0.337 $\pm$ 0.009 & 0.967 $\pm$ 0.013 & 48.486 $\pm$ 0.042 & 56.897 $\pm$ 0.059 & 1.135 $\pm$ 0.004 \\
\midrule
\multirow{2}{*}{300K} 
  & \nameshort & \textbf{0.112 $\pm$ 0.006} & \textbf{0.379 $\pm$ 0.028} & \textbf{1.530 $\pm$ 0.068} & \textbf{1.615 $\pm$ 0.053} & \textbf{0.270 $\pm$ 0.023} \\
  & TA-BG      & 0.219 $\pm$ 0.013 & 0.685 $\pm$ 0.034 & 83.457 $\pm$ 0.070 & 86.176 $\pm$ 0.104 & 0.979 $\pm$ 0.012 \\
\bottomrule
\end{tabular}
}
\label{tab:aldp_temperatures}
\end{table}

\section{Training Dynamics Across Temperatures}
\label{app:analysis_temperature}
\begin{figure}[htbp]
  \centering
  \includegraphics[width=\linewidth]{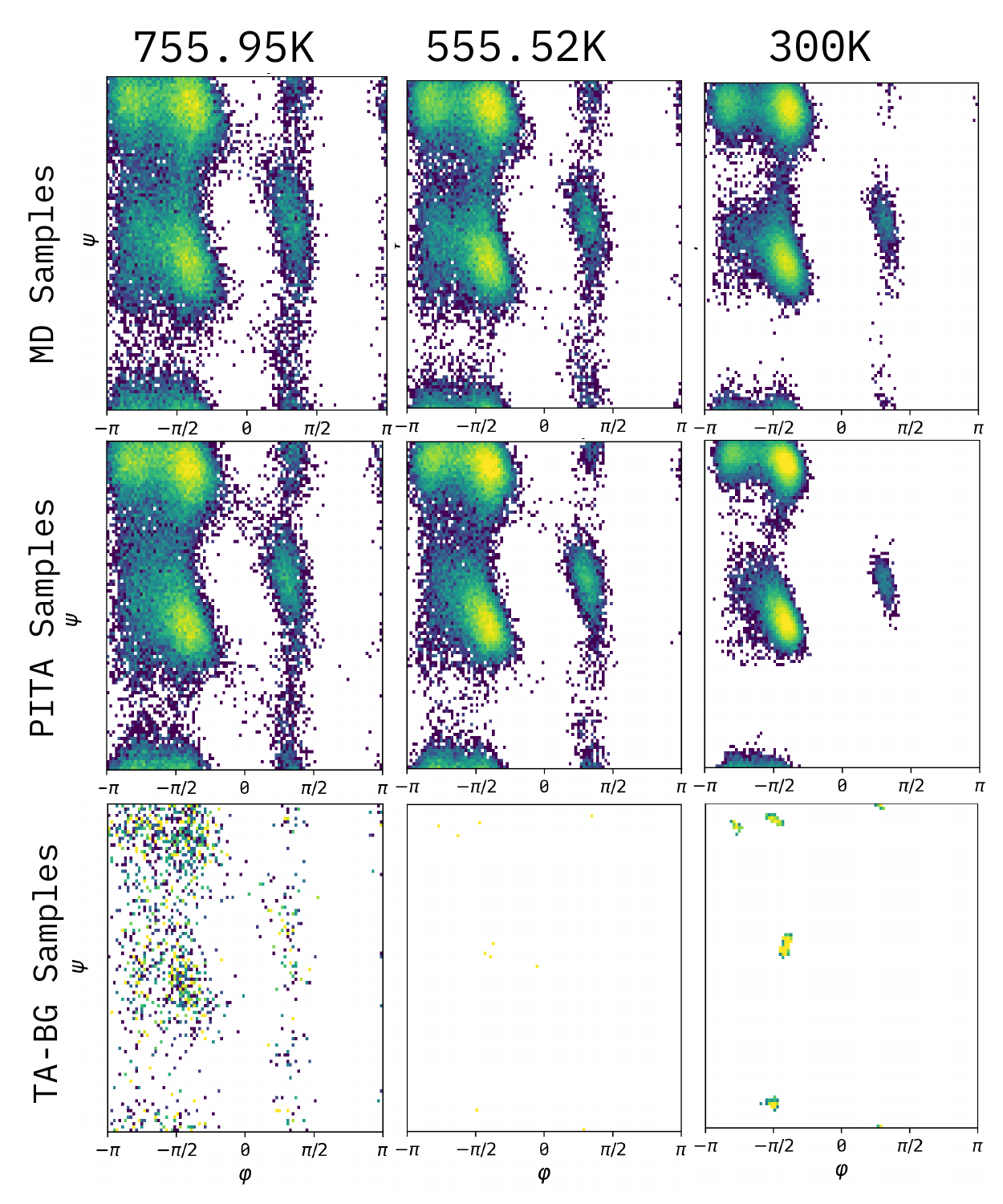}
  \caption{Ramachandran plots for Alanine Dipeptide (ALDP) obtained from different temperatures using 30k samples. We compare the samples from \nameshort and TA-BG with the ground-truth MD samples.}
  \label{fig:all_temps_aldp}
\end{figure}

In this section, we analyze the performance of the models (\nameshort and TA-BG) across different temperatures during annealing toward the target temperature on ALDP. Table~\ref{tab:aldp_temperatures} presents quantitative metrics, demonstrating that \nameshort consistently achieves lower discrepancies across all temperatures. Additionally, Figure~\ref{fig:all_temps_aldp} shows the Ramachandran plots at temperatures, further illustrating the ability of the model to generate physically realistic samples that capture the temperature-dependent conformational landscape at each step of the annealing process. TA-BG demonstrates reasonable performance at $755.95$K when initialized with ground-truth samples, reflecting its ability to model high-temperature distributions under ideal conditions. However, its performance deteriorates when transitioning to lower temperatures using recursively generated samples for importance sampling, indicated by the mode collapse in the Ramachandran plots, where the conformational diversity sharply diminishes. 

\section{Additional Experimental Details}
\label{app:experiment_details}
\subsection{Parameterization Details}

\xhdr{\nameshort}
For LJ-13, we use equal loss weights for energy pinning, denoising score matching, and EBM distillation. We use the noise schedule of \citet{karras2022elucidating}, with the following parameters: $\sigma_{\mathrm{min}}=0.05$, $\sigma_{\mathrm{max}}=80$ and $\rho=7$. The model uses EGNN~\citep{satorras2022enequivariantgraphneural} with approximately 90k parameters, consisting of three layers and a hidden dimension of 32. For ALDP and AL3, the energy pinning, denoising score matching, and EBM distillation components of the loss are weighted equally at 1.0, with an additional target score matching loss weighted at 0.01. We use the same noise schedule as the LJ-13 experiment, using a smaller $\sigma_\mathrm{min}$ of 0.01. We use DiT~\citep{peebles2023scalablediffusionmodelstransformers} comprising six layers and six attention heads, with a hidden size of 192 and a total of roughly 12 million parameters. All models are trained with a learning rate of $1 \times 10^{-3}$ without any weight decay. For ALDP and AL3, we use Exponential Moving Average (EMA) with a decay rate of 0.999, updating every gradient step. 

\xhdr{MD-Diff}
We train the diffusion model on the MD trajectories generated at the target temperature. This serves as a strong baseline, since we have direct access to the ground-truth samples, unlike \nameshort and TA-BG. To ensure a controlled comparison, the length of the MD chain used to train the diffusion model is chosen such that the total number of energy evaluations matches the computational budget used to train \nameshort over all annealing steps. We provide further analysis on the mixing properties of different lengths of MD chains at low and high temperatures in \cref{app:md-mixing}. We use a $\sigma_\mathrm{min}$ value of 0.005, while keeping the rest of the model hyperparameters the same as \nameshort. 

\looseness=-1
\xhdr{TA-BG}
In~\citet{schopmans2025temperatureannealedboltzmanngenerators}, TA-BG trains a normalizing flow by minimizing the reverse Kullback–Leibler (KL) divergence at high temperature and progressively refining the model via importance sampling as the temperature is annealed toward the target distribution. We carefully adapt their training pipeline to ensure a consistent and fair comparison. Specifically, we initialize the training with ground-truth MD data rather than learned high-temperature samples, represent molecular configurations in Cartesian rather than internal coordinates, and use the same temperature annealing schedule as \nameshort. We use TarFlow \citep{zhai2024normalizingflowscapablegenerative}, configured with four meta blocks, each containing four attention layers and a hidden size of 256, resulting in approximately 12 million parameters. We use a learning rate of $1 \times 10^{-4}$ and employ 60,000 samples at the end of training for each temperature to compute the importance weights used in generating the buffer for the next temperature.

\xhdr{MD-NF}
Similarly to MD-Diff, we train the normalizing flow model on the MD trajectories generated at the target temperature. We use TarFlow with the same model hyperparameters that we used for training TA-BG.

\xhdr{Score Scaling} We use a Score Scaling baseline which is a simple (but biased) modification to the score to attempt to sample from a different temperature. Specifically, given a score function $s_t(x_t)$ where we would normally sample with the SDE
\begin{equation}
    d x_t = \left ( - a_t x_t + \zeta_t^2 \frac{1+\xi_t}{2} s_t(x_t) \right ) dt + 
    \zeta_t \sqrt{\xi_t} d W_t, \quad x_0 \sim q_{t=0}(x)
\end{equation}
we instead sample with 
\begin{equation}
    d x_t = \left ( - a_t x_t + \gamma \zeta_t^2 \frac{1+\xi_t}{2} s_t(x_t) \right ) dt + \zeta_t \sqrt{\xi_t} d W_t, \quad x_0 \sim q_{t=0}(x)
\end{equation}

\subsection{Metrics}
\label{app:metrics}
We evaluate model performance using both sample-based metrics and metrics that assess energy distributions. To compare energy distributions between generated samples and ground-truth molecular dynamics (MD) samples, we compute the 1D 1-Wasserstein and 2-Wasserstein distances on the energy histograms. For sample-based evaluation, we measure the 2D wrapped 2-Wasserstein distance of the internal dihedral angles, $\phi$ and $\psi$ (denoted as $\mathbb{T}$-$\mathcal{W}_2$). Additionally, we calculate the 2D 1-Wasserstein and 2-Wasserstein distances between the first two TICA axes of the ground-truth and generated samples.

\subsection{MD Parameters}

\looseness=-1
\xhdr{LJ-13 Parameters} 
The Lennard-Jones (LJ) potential is an intermolecular potential that models interactions of non-bonding particles. The energy is a function of the interatomic distance of the particles:
\begin{equation}\label{eq:lj}
    \gE^{\rm LJ}(x) = \frac{\eps}{2 \tau} \sum_{ij} \left( \left( \frac{r_m}{d_{ij}}\right)^6 - \left( \frac{r_m}{d_{ij}}\right)^{12} \right)
\end{equation}
where the distance between two particles $i$ and $j$ is $d_{ij}=\|x_i - x_j\|_2$ and $r_m$, $c$, $\epsilon$ and $c_{osc}$ are physical constants. As in \citet{kohler2020equivariant}, we also add a harmonic potential to the energy so that $\gE^{LJ-system} = \gE^{\rm LJ}(x) + c_{osc} \gE^{\rm osc}(x)$
This harmonic potential is given by:
\begin{equation}
    \gE^{\rm osc}(x) = \frac{1}{2} \sum_{i}||x_i - x_{\rm COM}||^2
\end{equation}
where $x_{\rm COM}$ is the center of mass of the system. 
We use $r_m=1$, $c=1$, $\eps=2.0$ and $c = 1.0$. 
For the LJ-13 dataset, we draw MCMC chains using the No-U-Turn-Sampler (NUTS)~\citep{hoffman2011nouturnsampleradaptivelysetting}

\xhdr{Alanine Parameters} For MD data on ALDP and AL,3, we run two chains one for training and one for test. We use the same simulation parameters for both. For training data we sample shorter chains more frequently (every 100 md steps). To conserve disk space for long test chains, we save every 10k steps. Further parameters can be found in \cref{tab:md_parameters} and \cref{tab:dataset_parameters}. 

\begin{table}[h]
\centering
\caption{\texttt{OpenMM} simulation parameters.}
\label{tab:md_parameters}
\begin{tabular}{ll}
\toprule
Force field         & \texttt{amber-14} \\
Integration time step            & 1 fs \\
Friction coefficient & \SI{0.3}{\ps}\(^{-1}\) \\
Temperature          & \SI{300}{\K} \\
Nonbonded method      & \texttt{CutoffNonPeriodic} \\
Nonbonded cutoff      & \SI{2}{\nm} \\
Integrator           & \texttt{LangevinMiddleIntegrator} \\
\bottomrule
\end{tabular}
\end{table}

\begin{table}[h]
\centering
\caption{Training and evaluation dataset parameters.}
\label{tab:dataset_parameters}
\begin{tabular}{lll}
\toprule
 & Train & Test \\
 \midrule
Burn-in period & \SI{50}{\ps} & \SI{50}{\ps} \\
Sampling interval & \SI{0.1}{\ps} & \SI{10}{\ps} \\
Simulation time & \SI{50}{\ns} & \SI{1}{\us} \\
\bottomrule
\end{tabular}
\end{table}

\section{Pseudocode}
In this section we provide Python pseudocode for easy of understanding and reimplementation.

\begin{lstlisting}[language=Python, caption={Python implementation of resampled inference.}, label={lst:resampled_inference}]
def resampled_inference(x0, T, score_model, energy_model, gamma, a, zeta, xi):
    xt = x0
    dt = 1 / T
    for t in linspace(0,1,T+1)[:-1]:
        # Define variables
        st = score_model(xt, t)
        ut = energy_model(xt, t)
        grad_Ut = grad(ut, xt)
        dUt_dt = grad(ut, t)
        # Equation (11)
        drift = (-a(t) * xt) + zeta(t)**2 / 2 * (st - gamma * xi(t) * grad_Ut)
        diffusion = zeta(t) * sqrt(xi(t)) * randn_like(xt)
        xt += drift * dt + diffusion * sqrt(dt)
        # Equation (12)
        dA = div(xt) * zeta(t) ** 2/ 2 - gamma * grad_Ut * ((-a(t) * xt) + zeta(t) ** 2 / 2 * st) - gamma * dU_dt
        At = dA * dt
        # Resample xt proportional to At with quasi monte carlo
        xt = resample(xt, At)
    return xt
\end{lstlisting}

\section{Extended Related Work}

\looseness=-1
\xhdr{Annealed Importance Sampling} In the context of AIS \citep{jarzynski1997nonequilibrium, neal2001annealed}, SMC samplers \citep{Del-Moral:2006} and parallel tempering \citep{swendsen1986replica}, our method reduces the number of energy evaluations by learning the models of intermediate marginals. Indeed, when the buffer of samples from the current temperature is sampled, training of the diffusion model does not require new energy evaluations (note that the gradients for target score matching can be cached). Thus, the only time we need to evaluate the energies is for the importance sampling at the final step of the inference-time annealing and for the collection of samples via MCMC at a high temperature. Obviously, for sampling from the target density $\pi(x)$, the trained diffusion model, unlike AIS, allows producing uncorrelated samples without restarting the chain from the prior distribution.

%% file: main.bbl
\begin{thebibliography}{}

\bibitem[Akhound-Sadegh et~al., 2024]{akhound2024iterated}
Akhound-Sadegh, T., Rector-Brooks, J., Bose, A.~J., Mittal, S., Lemos, P., Liu, C.-H., Sendera, M., Ravanbakhsh, S., Gidel, G., Bengio, Y., Malkin, N., and Tong, A. (2024).
\newblock Iterated denoising energy matching for sampling from {B}oltzmann densities.
\newblock In {\em International Conference on Machine Learning}.

\bibitem[Albergo and Vanden-Eijnden, 2025]{albergo2024nets}
Albergo, M.~S. and Vanden-Eijnden, E. (2025).
\newblock Nets: A non-equilibrium transport sampler.
\newblock In {\em International Conference on Machine Learning}.

\bibitem[Berner et~al., 2024]{berner2022optimal}
Berner, J., Richter, L., and Ullrich, K. (2024).
\newblock An optimal control perspective on diffusion-based generative modeling.
\newblock {\em Transactions on Machine Learning Research}.

\bibitem[Blessing et~al., 2024]{blessing2024beyond}
Blessing, D., Jia, X., Esslinger, J., Vargas, F., and Neumann, G. (2024).
\newblock Beyond {ELBO}s: A large-scale evaluation of variational methods for sampling.
\newblock In {\em International Conference on Machine Learning}.

\bibitem[Chen et~al., 2025]{chen2024sequential}
Chen, J., Richter, L., Berner, J., Blessing, D., Neumann, G., and Anandkumar, A. (2025).
\newblock Sequential controlled {L}angevin diffusions.
\newblock In {\em International Conference on Learning Representations}.

\bibitem[Chen et~al., 2018]{chen2018neural}
Chen, R.~T., Rubanova, Y., Bettencourt, J., and Duvenaud, D.~K. (2018).
\newblock Neural ordinary differential equations.
\newblock In {\em Neural Information Processing Systems}.

\bibitem[Chen et~al., 2023]{chen2023sampling}
Chen, S., Chewi, S., Li, J., Li, Y., Salim, A., and Zhang, A.~R. (2023).
\newblock Sampling is as easy as learning the score: theory for diffusion models with minimal data assumptions.
\newblock In {\em International Conference on Learning Representations}.

\bibitem[De~Bortoli et~al., 2024]{de2024target}
De~Bortoli, V., Hutchinson, M., Wirnsberger, P., and Doucet, A. (2024).
\newblock Target score matching.
\newblock {\em arXiv preprint arXiv:2402.08667}.

\bibitem[Del~Moral et~al., 2006]{Del-Moral:2006}
Del~Moral, P., Doucet, A., and Jasra, A. (2006).
\newblock Sequential {M}onte {C}arlo samplers.
\newblock {\em Journal of the Royal Statistical Society: Series B}, 68(3):411--436.

\bibitem[Du et~al., 2023]{du2023reduce}
Du, Y., Durkan, C., Strudel, R., Tenenbaum, J.~B., Dieleman, S., Fergus, R., Sohl-Dickstein, J., Doucet, A., and Grathwohl, W.~S. (2023).
\newblock Reduce, reuse, recycle: Compositional generation with energy-based diffusion models and {MCMC}.
\newblock In {\em International Conference on Machine Learning}.

\bibitem[Frenkel and Smit, 2023]{frenkel2023understanding}
Frenkel, D. and Smit, B. (2023).
\newblock {\em Understanding Molecular Simulation: from Algorithms to Applications}.
\newblock Elsevier.

\bibitem[Havens et~al., 2025]{havens2025adjoint}
Havens, A., Miller, B.~K., Yan, B., Domingo-Enrich, C., Sriram, A., Wood, B., Levine, D., Hu, B., Amos, B., Karrer, B., Fu, X., Liu, G.-H., and Chen, R. (2025).
\newblock Adjoint sampling: Highly scalable diffusion samplers via adjoint matching.
\newblock In {\em International Conference on Machine Learning}.

\bibitem[He et~al., 2025]{he2025no}
He, J., Du, Y., Vargas, F., Zhang, D., Padhy, S., OuYang, R., Gomes, C., and Hern{\'a}ndez-Lobato, J.~M. (2025).
\newblock No trick, no treat: Pursuits and challenges towards simulation-free training of neural samplers.
\newblock {\em arXiv preprint arXiv:2502.06685}.

\bibitem[H{\'e}nin et~al., 2022]{henin2022enhanced}
H{\'e}nin, J., Leli{\`e}vre, T., Shirts, M.~R., Valsson, O., and Delemotte, L. (2022).
\newblock Enhanced sampling methods for molecular dynamics simulations.
\newblock {\em Living Journal of Computational Molecular Science}, 4(1).

\bibitem[Ho et~al., 2020]{ho2020denoising}
Ho, J., Jain, A., and Abbeel, P. (2020).
\newblock Denoising diffusion probabilistic models.
\newblock In {\em Neural Information Processing Systems}.

\bibitem[Hoffman and Gelman, 2014]{hoffman2011nouturnsampleradaptivelysetting}
Hoffman, M.~D. and Gelman, A. (2014).
\newblock The {No-U-Turn} sampler: Adaptively setting path lengths in {H}amiltonian {M}onte {C}arlo.
\newblock {\em Journal of Machine Learning Research}.

\bibitem[Huang et~al., 2025]{huang2021schrodinger}
Huang, J., Jiao, Y., Kang, L., Liao, X., Liu, J., and Liu, Y. (2025).
\newblock Schr{\"o}dinger-{F}{\"o}llmer sampler: sampling without ergodicity.
\newblock {\em IEEE Transactions on Information Theory}.

\bibitem[Jarzynski, 1997]{jarzynski1997nonequilibrium}
Jarzynski, C. (1997).
\newblock Nonequilibrium equality for free energy differences.
\newblock {\em Physical Review Letters}, 78(14):2690--2963.

\bibitem[Karras et~al., 2022]{karras2022elucidating}
Karras, T., Aittala, M., Aila, T., and Laine, S. (2022).
\newblock Elucidating the design space of diffusion-based generative models.
\newblock In {\em Neural Information Processing Systems}.

\bibitem[Klein et~al., 2023]{klein2023equivariant}
Klein, L., Kr{\"a}mer, A., and No{\'e}, F. (2023).
\newblock Equivariant flow matching.
\newblock In {\em Neural Information Processing Systems}.

\bibitem[Klein and Noé, 2024]{klein2024transferableboltzmanngenerators}
Klein, L. and Noé, F. (2024).
\newblock Transferable {B}oltzmann generators.
\newblock In {\em Neural Information Processing Systems}.

\bibitem[K{\"o}hler et~al., 2020]{kohler2020equivariant}
K{\"o}hler, J., Klein, L., and No{\'e}, F. (2020).
\newblock Equivariant flows: exact likelihood generative learning for symmetric densities.
\newblock In {\em International Conference on Machine Learning}.

\bibitem[Leimkuhler and Matthews, 2015]{leimkuhler2015molecular}
Leimkuhler, B. and Matthews, C. (2015).
\newblock {\em Molecular Dynamics: With Deterministic and Stochastic Numerical Methods}.
\newblock Springer.

\bibitem[Lipman et~al., 2023]{lipman2022flow}
Lipman, Y., Chen, R.~T., Ben-Hamu, H., Nickel, M., and Le, M. (2023).
\newblock Flow matching for generative modeling.
\newblock In {\em International Conference on Learning Representations}.

\bibitem[Liu, 2001]{liu2001monte}
Liu, J.~S. (2001).
\newblock {\em Monte Carlo Strategies in Scientific Computing}.
\newblock Springer.

\bibitem[Neal, 2001]{neal2001annealed}
Neal, R.~M. (2001).
\newblock Annealed importance sampling.
\newblock {\em Statistics and Computing}, 11(2):125--139.

\bibitem[Neklyudov et~al., 2023]{neklyudov2023action}
Neklyudov, K., Brekelmans, R., Severo, D., and Makhzani, A. (2023).
\newblock Action matching: Learning stochastic dynamics from samples.
\newblock In {\em International Conference on Machine Learning}.

\bibitem[Noble et~al., 2025]{noble2025learned}
Noble, M., Grenioux, L., Gabrié, M., and Durmus, A.~O. (2025).
\newblock Learned reference-based diffusion sampling for multi-modal distributions.
\newblock In {\em International Conference on Learning Representations}.

\bibitem[No{\'e} et~al., 2019]{noe2019boltzmann}
No{\'e}, F., Olsson, S., K{\"o}hler, J., and Wu, H. (2019).
\newblock Boltzmann generators: Sampling equilibrium states of many-body systems with deep learning.
\newblock {\em Science}, 365(6457):eaaw1147.

\bibitem[Ohno et~al., 2018]{ohno2018computational}
Ohno, K., Esfarjani, K., and Kawazoe, Y. (2018).
\newblock {\em Computational Materials Science: From Ab Initio to {M}onte {C}arlo Methods}.
\newblock Springer.

\bibitem[Peebles and Xie, 2023]{peebles2023scalablediffusionmodelstransformers}
Peebles, W. and Xie, S. (2023).
\newblock Scalable diffusion models with transformers.
\newblock In {\em Proceedings of the IEEE/CVF International Conference on Computer Vision}.

\bibitem[Richter et~al., 2024]{richter2023improved}
Richter, L., Berner, J., and Liu, G.-H. (2024).
\newblock Improved sampling via learned diffusions.
\newblock {\em International Conference on Learning Representations}.

\bibitem[Rissanen et~al., 2025]{rissanen2025progressivetemperingsamplerdiffusion}
Rissanen, S., OuYang, R., He, J., Chen, W., Heinonen, M., Solin, A., and Hernández-Lobato, J.~M. (2025).
\newblock Progressive tempering sampler with diffusion.
\newblock In {\em International Conference on Machine Learning}.

\bibitem[Satorras et~al., 2021]{satorras2022enequivariantgraphneural}
Satorras, V.~G., Hoogeboom, E., and Welling, M. (2021).
\newblock E (n) equivariant graph neural networks.
\newblock In {\em International Conference on Machine Learning}.

\bibitem[Schopmans and Friederich, 2025]{schopmans2025temperatureannealedboltzmanngenerators}
Schopmans, H. and Friederich, P. (2025).
\newblock Temperature-annealed {B}oltzmann generators.
\newblock In {\em International Conference on Machine Learning}.

\bibitem[Skreta et~al., 2025]{skreta2025feynman}
Skreta, M., Akhound-Sadegh, T., Ohanesian, V., Bondesan, R., Aspuru-Guzik, A., Doucet, A., Brekelmans, R., Tong, A., and Neklyudov, K. (2025).
\newblock {Feynman--Kac} correctors in diffusion: Annealing, guidance, and product of experts.
\newblock In {\em International Conference on Machine Learning}.

\bibitem[Song et~al., 2021]{song2020score}
Song, Y., Sohl-Dickstein, J., Kingma, D.~P., Kumar, A., Ermon, S., and Poole, B. (2021).
\newblock Score-based generative modeling through stochastic differential equations.
\newblock In {\em International Conference on Learning Representations}.

\bibitem[Stoltz et~al., 2010]{stoltz2010free}
Stoltz, G., Rousset, M., and Leli\`evre, T. (2010).
\newblock {\em Free Energy Computations: A Mathematical Perspective}.
\newblock World Scientific.

\bibitem[Swendsen and Wang, 1986]{swendsen1986replica}
Swendsen, R.~H. and Wang, J.-S. (1986).
\newblock Replica {M}onte {C}arlo simulation of spin-glasses.
\newblock {\em Physical Review Letters}, 57(21):2607.

\bibitem[Tan et~al., 2025a]{tan2025scalable}
Tan, C.~B., Bose, A.~J., Lin, C., Klein, L., Bronstein, M.~M., and Tong, A. (2025a).
\newblock Scalable equilibrium sampling with sequential {B}oltzmann generators.
\newblock In {\em International Conference on Machine Learning}.

\bibitem[Tan et~al., 2025b]{tan2025amortizedsamplingtransferablenormalizing}
Tan, C.~B., Hassan, M., Klein, L., Syed, S., Beaini, D., Bronstein, M.~M., Tong, A., and Neklyudov, K. (2025b).
\newblock Amortized sampling with transferable normalizing flows.
\newblock In {\em Neural Information Processing Systems}.

\bibitem[Thornton et~al., 2025]{thornton2025composition}
Thornton, J., Bethune, L., Zhang, R., Bradley, A., Nakkiran, P., and Zhai, S. (2025).
\newblock Composition and control with distilled energy diffusion models and sequential {M}onte {C}arlo.
\newblock In {\em International Conference on Artificial Intelligence and Statistics}.

\bibitem[Vaikuntanathan and Jarzynski, 2008]{vaikuntanathan2008escorted}
Vaikuntanathan, S. and Jarzynski, C. (2008).
\newblock Escorted free energy simulations: Improving convergence by reducing dissipation.
\newblock {\em Physical Review Letters}, 100(19):190601.

\bibitem[Vargas et~al., 2023]{vargasdenoising}
Vargas, F., Grathwohl, W.~S., and Doucet, A. (2023).
\newblock Denoising diffusion samplers.
\newblock In {\em International Conference on Learning Representations}.

\bibitem[Vargas et~al., 2024]{vargas2024transport}
Vargas, F., Padhy, S., Blessing, D., and N{\"u}sken, N. (2024).
\newblock Transport meets variational inference: Controlled {Monte Carlo} diffusions.
\newblock In {\em International Conference on Learning Representations}.

\bibitem[Woodard et~al., 2009]{woodard2009sufficient}
Woodard, D., Schmidler, S., and Huber, M. (2009).
\newblock Sufficient conditions for torpid mixing of parallel and simulated tempering.
\newblock {\em Electronic Journal of Probability}.

\bibitem[Zhai et~al., 2025]{zhai2024normalizingflowscapablegenerative}
Zhai, S., Zhang, R., Nakkiran, P., Berthelot, D., Gu, J., Zheng, H., Chen, T., Bautista, M.~A., Jaitly, N., and Susskind, J. (2025).
\newblock Normalizing flows are capable generative models.
\newblock In {\em International Conference on Machine Learning}.

\bibitem[Zhang and Chen, 2022]{zhangyongxinchen2021path}
Zhang, Q. and Chen, Y. (2022).
\newblock Path integral sampler: a stochastic control approach for sampling.
\newblock In {\em International Conference on Learning Representations}.

\end{thebibliography}
